\documentclass[preprint,12pt,review,3p]{elsarticle}

\usepackage{amssymb, amsmath, bm, amsfonts}
\usepackage{siunitx}
\usepackage{utfsym}
\renewcommand{\checkmark}{\usym{2713}}
\newcommand{\xmark}{\usym{2717}}
\usepackage{float,graphbox}
\usepackage{epstopdf}
\usepackage{subcaption}
\usepackage{hyperref}
\usepackage{color, xcolor}
\usepackage{booktabs, multirow, makecell}
\usepackage{threeparttable}
\usepackage{array,tabularray}
\usepackage{titlesec}
\usepackage[ruled,linesnumbered]{algorithm2e}
\usepackage{enumitem}
\setlist{nosep}
\usepackage{soul}
\usepackage{lipsum}

\journal{Elsevier}

\begin{document}
\begin{frontmatter}


    \title{Sharp-PINNs: staggered hard-constrained physics-informed neural networks for phase field modelling of corrosion}

    \author{Nanxi Chen\fnref{TJU}}
    \author{Chuanjie Cui\fnref{Oxford}}
    \author{Rujin Ma\fnref{TJU}\corref{cor1}}\ead{rjma@tongji.edu.cn}
    \author{Airong Chen\fnref{TJU}}
    \author{Sifan Wang\fnref{Yale}\corref{cor1}}\ead{sifan.wang@yale.edu}

    \address[TJU]{College of Civil Engineering, Tongji University, Shanghai 200092, China}
    \address[Oxford]{Department of Engineering Science, University of Oxford, Oxford OX1 3PJ, UK}
    \address[Yale]{Institution for Foundation of Data Science, Yale University, New Haven, CT 06520, USA}

    \cortext[cor1]{Corresponding author}





    \begin{abstract}
        Physics-informed neural networks have shown significant potential in solving partial differential equations (PDEs) across diverse scientific fields. However, their performance often deteriorates when addressing PDEs with intricate and strongly coupled solutions. In this work, we present a novel Sharp-PINN framework to tackle complex phase field corrosion problems. Instead of minimizing all governing PDE residuals simultaneously, the Sharp-PINNs introduce a staggered training scheme that alternately minimizes the residuals of Allen-Cahn and Cahn-Hilliard equations, which govern the corrosion system. To further enhance its efficiency and accuracy, we design an advanced neural network architecture that integrates random Fourier features as coordinate embeddings, employs a modified multi-layer perceptron as the primary backbone, and enforces hard constraints in the output layer. This framework is benchmarked through simulations of corrosion problems with multiple pits, where the staggered training scheme and network architecture significantly improve both the efficiency and accuracy of PINNs. Moreover, in three-dimensional cases, our approach is 5-10 times faster than traditional finite element methods while maintaining competitive accuracy, demonstrating its potential for real-world engineering applications in corrosion prediction.
    \end{abstract}


\end{frontmatter}

\section{Introduction}
\label{sec:introduction}

Corrosion is a pervasive yet destructive issue in many engineering structures and systems, leading to material degradation and potentially severe failures \cite{guoCriticalReviewCorrosion2019,fesslerPipelineCorrosion2008,coleSciencePipeCorrosion2012,soaresInfluenceEnvironmentalFactors2009}. This phenomenon poses considerable economic and safety challenges, impacting both public safety and infrastructure durability. Therefore, a comprehensive understanding of the fundamental mechanisms of corrosion, along with the development of advanced computational models for its prediction and mitigation, remains a critical research priority within the scientific and engineering community \cite{marcusCorrosionMechanismsTheory2002}.

Corrosion is a complex multi-physics problem that involves the interplay of electrochemical reactions, material dissolution, and mass transport across different phases \cite{jafarzadehComputationalModelingPitting2019}, and even mechanical effects when stress corrosion cracking is evolved \cite{Turnbull2001,JMPS2021}. Thus, the development of reliable predictive models is often hindered by the strong coupling of these physical processes, both theoretically and numerically. Another major challenge in modelling corrosion lies in its discontinuous nature---specifically, how the sharp metal–electrolyte interface is defined and evolves over time.  Moreover, localized corrosion, such as pitting, presents additional difficulties due to the merging of numerous pits within a confined region. The interplay and coalescence of these pits further complicate the problem. Accurately capturing complex corrosion morphologies in numerical models often requires handling moving interfacial boundary conditions and manually adjusting interface topology based on predefined criteria. Existing approaches include the eXtended Finite Element Method (X-FEM) \cite{Duddu2016}, Arbitrary Lagrangian-Eulerian (ALE) \cite{ArbitraryLagrangianEulerian2014}, and Level Set (LS) techniques \cite{dudduNumericalModelingCorrosion2014}. However, these techniques often struggle when handling complex coalescence processes and three-dimensional geometries. An increasingly popular alternative is the phase field method, which transforms sharp interface into a \textit{diffused} region. By reformulating complex interfacial problems into the solution of Partial Differential Equations (PDEs)---such as the Allen-Cahn (AC) equation for non-conserved variables and the Cahn-Hilliard (CH) equation for conserved variables \cite{cahnFreeEnergyNonuniform1958, allenGroundStateStructures1972}---the phase field approach eliminates the need for explicit boundary tracking and allows the evolution of moving interfaces to emerge naturally as a result of the governing equations \cite{Biner2017}. This enables seamless simulation of dynamic interface evolution in various materials science applications. Furthermore, the governing equations of the phase field method are derived from the total system energy using thermodynamic principles, making it particularly well-suited for multi-physics coupling problems. Numerous studies have demonstrated the effectiveness of phase field models in simulating corrosion processes under different conditions \cite{Mai2016,Ansari2018,JMPS2021,JMPS2022,EWC2023}, highlighting its exceptional capability in capturing complex corrosion phenomena.

While phase field method has achieved remarkable success in the corrosion community, its practical adoption is often limited by significant computational challenges. Traditional numerical framework, such as finite element method, struggle with the inherent nonlinearity and strong coupling of the AC and CH equations, leading to high computational costs and convergence difficulties \cite{keyesMultiphysicsSimulations2013, gomezReviewComputationalModelling2019}. Moreover, as spatial dimensionality increases, the exponential growth in degrees of freedom—commonly known as the \textit{curse of dimensionality}, makes large scale simulations prohibitively resource intensive \cite{hughesFiniteElementMethod2003, JMPS2021}. These factors drive the academic community to continuously seek alternative numerical approaches \cite{Ansari2021}. Physics-informed neural networks (PINNs) offer a transformative approach to solving PDEs by embedding governing equations and boundary conditions directly into a neural network's loss function. This mesh-free paradigm removes the need for complex grid generation and numerical stabilization techniques, addressing common challenges in conventional methods \cite{raissiPhysicsinformedNeuralNetworks2019,karniadakisPhysicsinformedMachineLearning2021}. Additionally, PINNs mitigate the \textit{curse of dimensionality} as their number of collocation points and trainable parameters scales more efficiently with dimensionality, making them particularly suited for high-dimensional PDEs with intricate solutions \cite{TacklingCurseDimensionality2024}. All these distinctive advantages position PINNs as a promising alternative to traditional finite element methods, with applications spanning various scientific fields including fluid dynamics \cite{caiPhysicsinformedNeuralNetworks2021b,sharmaReviewPhysicsInformedMachine2023,zhaoComprehensiveReviewAdvances2024}, solid mechanics \cite{raoPhysicsInformedDeepLearning2021,goswamiTransferLearningEnhanced2019,zhengPhysicsinformedMachineLearning2022}, materials science \cite{punPhysicallyInformedArtificial2019a,zhangAnalysesInternalStructures2022a}, heat conduction/convection analysis \cite{caiPhysicsinformedNeuralNetworks2021e,laubscherSimulationMultispeciesFlow2021,zobeiryPhysicsinformedMachineLearning2021}, electro-magnetics \cite{khanPhysicsInformedNeural2022,baldanPhysicsinformedNeuralNetworks2023a}, and geophysics \cite{rasht-beheshtPhysicsInformedNeuralNetworks2022a,schusterReviewPhysicsinformedMachinelearning2024,liProbabilisticPhysicsinformedNeural2024}.

In recent years, implementing PINNs for phase field equations has emerged as an increasingly prominent research topic \cite{wight2020solving,qiuPhysicsinformedNeuralNetworks2022,matteyNovelSequentialMethod2022,zhangRobustPhysicsinformedNeural2023}, offering a new avenue for addressing interfacial problems. However, most of these efforts have been restricted to solving the AC or CH equations individually, with PINNs for coupled AC-CH phase field models remaining largely unexplored. To address this gap, the authors recently proposed a generalized PF-PINNs framework incorporating advanced sampling and weighting strategies to mitigate the convergence challenges associated with the coupled AC-CH phase field model \cite{chenPFPINNsPhysicsinformedNeural2025}. However, optimizing PINNs for coupled AC-CH equations, especially in the presence of complex interfacial coalescence, remains an ongoing challenge.

In this study, we propose a novel Sharp-PINN framework tailored to solve complex phase field corrosion problems, with a particular emphasis on the efficient and accurate simulation of multiple pits interactions and three-dimensional domains. The key contributions of this work are highlighted as follows:
\begin{itemize} 
    \item \textbf{A staggered training scheme} designed to address competitive optimization challenges, significantly enhancing the training efficiency for coupled AC and CH equations.
    \item \textbf{An advanced neural network architecture} that integrates a random Fourier feature embedding, a modified multi-layer perceptron (MLP), and hard constraints into the input, backbone, and output layers, respectively, ensuring high accuracy and stability.
    \item \textbf{Comprehensive benchmarks} validate the proposed Sharp-PINN framework against traditional finite element methods for solving complex phase field corrosion problems, demonstrating its superior computational efficiency in three-dimensional corrosion scenarios.
\end{itemize}

The remainder of this paper is organized as follows: Section~\ref{sec:basic-theory} provides a brief overview of PINNs formulation and the phase field corrosion model. Section~\ref{sec:methodology} presents the proposed Sharp-PINNs framework, including the staggered training scheme and advanced neural network architecture. Detailed numerical experiments and representative results are discussed in Section~\ref{sec:results-and-discussion}, validating the performance of the Sharp-PINNs in solving complex phase field corrosion problems. Finally, Section~\ref{sec:conclusion} concludes the paper with a summary of the key findings and future research directions.


\section{Basic theory of PINNs and phase field corrosion model}
\label{sec:basic-theory}

\subsection{Physics-informed neural networks (PINNs)}

PINNs are a class of deep learning models designed to solve partial differential equations (PDEs) by 
integrating neural networks with  physical priors during training process. This is achieved by embedding the residual of the governing equations directly into the loss function, enabling the network to learn the underlying physics in a soft manner \cite{raissiPhysicsinformedNeuralNetworks2019}.

Let us consider a physical system governed by a set of PDEs with input variables (spatial coordinates $\bm x$ and temporal coordinate $t$) and parameters $\bm \lambda$. The general form of the PDEs can be expressed as:
\begin{subequations}
    \begin{align}
        \mathcal{G}\left[\bm x, t, u(\bm x, t; \bm\lambda)\right] & = 0, \quad \bm x \in \mathcal{D},  \label{eq:pde-governing} \\
        \mathcal{B}\left[\bm x, t, u(\bm x, t; \bm\lambda)\right] & = 0, \quad \bm x \in \partial \mathcal{D},                \\
        \mathcal{I}\left[\bm x, u(\bm x, 0; \bm\lambda)\right]    & = 0, \quad \bm x \in \mathcal{D},
    \end{align}
    \label{eq:pde}
\end{subequations}%
where $\mathcal{G}$, $\mathcal{B}$, and $\mathcal{I}$ denote the governing equations, boundary and initial conditions, respectively, $u(\bm x, t; \bm\lambda)$ denotes the spatial-temporal solution field, $\mathcal{D}$ and $\partial \mathcal{D}$ denote the spatial domain and the boundary, respectively.

The PINNs use a neural network $\mathcal{N}$ to approximate the solution field $u(\bm x, t; \bm\lambda)$, reads:
\begin{equation}
    \hat u(\bm x, t; \bm\lambda) = \mathcal{N}(\bm x, t; \bm \theta), \label{eq:nn-approximation}
\end{equation}%
where $\mathcal{N}$ is a deep neural network with trainable parameters $\bm \theta$. The network is trained to minimize the loss function $\mathcal{L}$, which is defined as the weighted sum of the mean squared error (MSE) of the residuals of the governing equations, boundary conditions, and initial conditions:
\begin{equation}
    \mathcal{L} = w_g\mathcal{L}_g + w_b\mathcal{L}_b + w_i\mathcal{L}_i, \label{eq:weighted-loss}
\end{equation}%
with
\begin{subequations}
    \begin{align}
        \mathcal{L}_g &= \frac{1}{N_g} \sum_{i=1}^{N_g} \left| \mathcal{G}\left(\bm x_g^{j}, t_g^{j}, \hat u\right) \right|^2, \label{eq:loss-governing} \\
        \mathcal{L}_b &= \frac{1}{N_b} \sum_{i=1}^{N_b} \left| \mathcal{B}\left(\bm x_b^{j}, t_b^{j}, \hat u\right) \right|^2, \label{eq:loss-boundary} \\
        \mathcal{L}_i &= \frac{1}{N_i} \sum_{i=1}^{N_i} \left| \mathcal{I}\left(\bm x_i^j, \hat u\right) \right|^2. \label{eq:loss-initial}
    \end{align}
    \label{eq:each-loss}
\end{subequations}
Here, $\{ \bm x_g^j, t_g^j \}_{j=1}^{N_g}$, $\{ \bm x_b^j, t_b^j \}_{j=1}^{N_b}$ and $\{ \bm x_i^j \}_{j=1}^{N_i}$ denote the spatial-temporal coordinates of the collocation points for the governing equations, boundary conditions, and initial conditions, respectively. These points should be strategically sampled from the spatial-temporal domain and the boundary \cite{wuComprehensiveStudyNonadaptive2023}. The weights $w_g$, $w_b$, and $w_i$ are used to balance the contributions of each loss term and can be either determined empirically beforehand or tuned automatically during training \cite{wangUnderstandingMitigatingGradient2021a,wangWhenWhyPINNs2022}.

\subsection{Phase field model of corrosion}
\label{sec:phase field-model-corrosion}

The phase field method is a powerful computational framework for simulating microstructure evolution, particularly in systems with complex interfacial dynamics. It employs scalar field variables to describe the distribution and evolution of different phases. Interfaces are implicitly represented as regions where these variables exhibit rapid transitions. These phase field variables can be classified into conserved and non-conserved types, governed by the CH and AC equations, respectively \cite{cahnFreeEnergyNonuniform1958,allenGroundStateStructures1972}.

The corrosion system is described using a single conserved variable $c$ (representing the concentration of a chemical species) and a non-conserved variable $\phi$ (describing phase transformation), expressed as \cite{Mai2016, JMPS2021}:
\begin{subequations}
    \begin{align}
        \text{Cahn-Hilliard: } & \frac{\partial c (\bm x, t)}{\partial t} = \nabla \cdot  M \nabla \frac{\delta \mathcal{F}}{\delta c}, \label{eq:cahn-hilliard} \\
        \text{Allen-Cahn: } & \frac{\partial \phi (\bm x, t)}{\partial t} = -L \frac{\delta \mathcal{F}}{\delta \phi}, \label{eq:allen-cahn}
    \end{align}
    \label{eq:phase field-equations}
\end{subequations}%
where $M$ and $L$ denote the diffusion mobility and the interface kinetics parameter, respectively, and $\mathcal{F}$ represents the system's free energy functional. 

The free energy functional $\mathcal{F}$ consists of the bulk free energy $\mathcal{F}_\text{bulk}$ and the interfacial energy $\mathcal{F}_\text{int}$ given by:
\begin{equation}
    \mathcal{F} = \mathcal{F}_\text{bulk} + \mathcal{F}_\text{int} = 
    \int \left[ f_\text{bulk}(c, \phi) + \frac{\alpha_\phi}{2} \left(\nabla\phi\right)^2 + \frac{\alpha_c}{2} \left(\nabla c\right)^2 \right] \, \text{d}V,\label{eq:free-energy-functional}
\end{equation}%
where $f_\text{bulk}(c, \phi)$ is the bulk free energy density, and $\alpha_c$ and $\alpha_\phi$ are gradient energy coefficients for the conserved and non-conserved variables, respectively. 

Note that in practice, considering one single gradient energy term, either $\alpha_c$ or $\alpha_\phi$, is already sufficient to qualitatively capture the energy contributions from the diffuse interface \cite{maiPhaseFieldModel2016}. Thus, for simplicity, we set $\alpha_c = 0$, leading to the simplified expression for the free energy density:
\begin{equation}
    f = f_\text{bulk}(c, \phi) + \frac{\alpha_\phi}{2} \left(\nabla\phi\right)^2, \label{eq:free-energy-density}
\end{equation}%
where $c$ is the normalized concentration field, defined as $c = {c_\text{real}} / c_\text{solid}$. Here, $c_\text{real}$ denotes the actual concentration, and $c_\text{solid}$ is the concentration in the solid phase (concentration of atoms).

We use the KKS (Kim-Kim-Suzuki) model \cite{kimPhasefieldModelBinary1999} to describe the phase field corrosion system. In this model, the normalized concentration field $c$ is treated as a mixture of solid and liquid phases, each characterized by distinct concentrations $c_\text{S}$ and $c_\text{L}$, but sharing the same chemical potential, reads:
\begin{subequations}
    \begin{gather}
        c = h(\phi) c_\text{S} + \left[1 - h(\phi)\right] c_\text{L}, \label{eq:concentration-mixture}\\
        \frac{\partial f_\text{S}(c_\text{S})}{\partial c_\text{S}} = \frac{\partial f_\text{L}(c_\text{L})}{\partial c_\text{L}}. \label{eq:chemical-potential}
    \end{gather}
\end{subequations}%
Here, $\phi$ and $c$ are normalized to the $[0, 1]$, with $0$ and $1$ representing the liquid and solid phases, respectively. $f_\text{S}(c_\text{S})$ and $f_\text{L}(c_\text{L})$ denote the free energy densities of the solid and liquid phases, respectively. $h(\cdot)$ is a $C^\infty$-continuous interpolation function that ensures a smooth transition between 0 and 1. In this study, the interpolation function is defined as:
\begin{equation}
    h(\phi) = -2\phi^3 + 3\phi^2.
\end{equation}

The local free energy density $f_\text{bulk}$ can be formulated using the weighted sum of the free energy densities of the solid and liquid phases:
\begin{equation}
    f_\text{bulk}(c, \phi) = h(\phi) f_\text{S}(c_\text{S}) + \left[1 - h(\phi)\right] f_\text{L}(c_\text{L}) + wg(\phi), \label{eq:bulk-free-energy}
\end{equation}%
where $w$ is the height of a double-well potential barrier and $g(\phi)=\phi^2(1-\phi)^2$. 

The formulations of the free energy densities associated with the solid and liquid phases are given by \cite{huThermodynamicDescriptionGrowth2007}:
\begin{subequations}
    \begin{gather}
        f_\text{S}(c_\text{S}) = \mathcal{A}\left(c_\text{S} - c_\text{Se}\right) ^2, \label{eq:solid-free-energy} \\ 
        f_\text{L}(c_\text{L}) = \mathcal{A}\left(c_\text{L} - c_\text{Le}\right) ^2. \label{eq:liquid-free-energy}
    \end{gather} \label{eq:free-energy-solid-liquid}
\end{subequations}%
Here, $\mathcal{A}$ is the curvature of the free energy density, $c_\text{Se} = c_\text{solid} / c_\text{solid}$ and $c_\text{Le} = c_\text{sat} / c_\text{solid}$ are the normalized equilibrium concentrations of the solid and liquid phases, respectively.
Taking Eqs.~\eqref{eq:solid-free-energy} and \eqref{eq:liquid-free-energy} into Eq.~\eqref{eq:chemical-potential}, yields the relationship between the equilibrium concentrations of the solid and liquid phases:
\begin{equation}
    c_\text{S} - c_\text{Se} = c_\text{L} - c_\text{Le}. \label{eq:equilibrium-concentration-relationship}
\end{equation}%
Inserting Eq. \eqref{eq:equilibrium-concentration-relationship} into Eq.~\eqref{eq:concentration-mixture}, the normalized concentration field of the solid phase $c_\text{S}$ and liquid phase $c_\text{L}$ can be expressed as:
\begin{subequations}
    \begin{align}
        c_\text{S} &= c + \left[ 1 - h(\phi) \right] \left( c_\text{Se} - c_\text{Le} \right), \label{eq:solid-concentration} \\
        c_\text{L} &= c - h(\phi) \left( c_\text{Se} - c_\text{Le} \right). \label{eq:liquid-concentration}
    \end{align}
    \label{eq:concentration-solid-liquid}
\end{subequations}%
Then, the free energy density $f$ can be rewritten as:
\begin{equation}
    f = \mathcal{A}\left[
        c - h(\phi)(c_\text{Se} - c_\text{Le}) - c_\text{Le}
    \right] ^2 + w_\phi g(\phi) + \frac{\alpha_\phi}{2} \left(\nabla\phi\right)^2. \label{eq:free-energy-density-simplified}
\end{equation}%
Finally, combining Eqs.~\eqref{eq:free-energy-density-simplified}, \eqref{eq:free-energy-functional} and \eqref{eq:phase field-equations}, the governing equations for the phase field corrosion model can be reformulated as:
\begin{subequations}
    \begin{align}
        \text{Cahn-Hilliard: } & \frac{\partial c}{\partial t}- 2\mathcal{A}M \Delta c + 2\mathcal{A}M \left(c_{\mathrm{Se}}-c_{\mathrm{Le}}\right) \Delta h\left(\phi\right) =0  \label{eq:chcorro} \\
        \text{Allen-Cahn: }    & \frac{\partial \phi}{\partial t}
        -2\mathcal{A}L\left[c-h(\phi)\left(c_{\mathrm{Se}}-c_{\mathrm{Le}}\right)-c_{\mathrm{Le}}\right]\left(c_{\mathrm{Se}}-c_{\mathrm{Le}}\right) h^{\prime}(\phi) +L w_\phi g^{\prime}(\phi) - L \alpha_\phi \Delta \phi =0.
        \label{eq:accorro}
    \end{align}
    \label{eq:governing-equations-corrosion}
\end{subequations}

\section{Sharp-PINN framework for solving phase field model of corrosion}
\label{sec:methodology}

\begin{figure}[htbp]
    \vspace{-1.2cm}
    \centering
    \includegraphics[width=.8\textwidth]{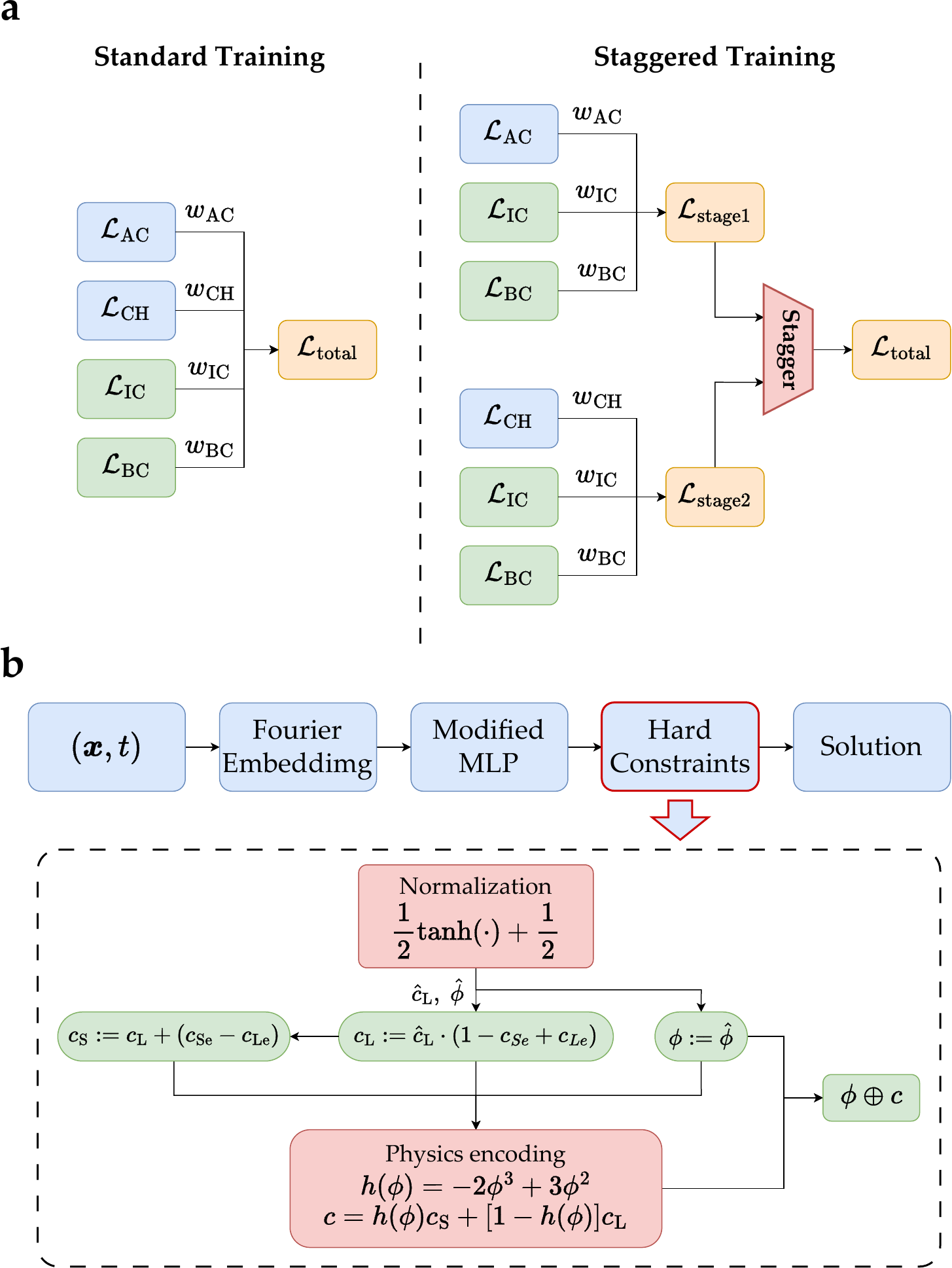}
    \caption{Schematic of the proposed Sharp-PINNs framework. (a) The staggered training scheme alternates between optimizing the Allen-Cahn (AC) and Cahn-Hilliard (CH) equations, effectively decoupling their optimization processes and mitigating competing gradients. Each training stage incorporates boundary conditions (BC) and initial conditions (IC) and the respective governing equation. (b) The proposed  neural network architecture consists of three key components: random Fourier feature ($\mathcal{F}$) for coordinate embedding, a modified MLP ($\mathcal{M}$) as the backbone, and a hard constraint layer ($\mathcal{H}$) that explicitly encodes the concentration field according to the KKS corrosion model.
    }
    \label{fig:proposed-pinn}
\end{figure}

In this section, we present the key innovations of Sharp-PINN framework for solving the phase field model of corrosion. Our framework features a novel staggered training scheme and an enhanced neural network architecture. Key features of this architecture include leveraging random Fourier features $\mathcal{F}$ as input embeddings, employing a modified MLP $\mathcal{M}$ as the primary backbone, and enforcing hard constraints $\mathcal{H}$ in the output layer. A schematic of the Sharp-PINN framework is shown in Figure~\ref{fig:proposed-pinn}.


\subsection{Staggered training scheme}
\label{sec:staggered-training}

The governing equations of the phase field corrosion model (AC and CH) are strongly coupled, reflecting the interplay between interfacial dynamics and mass conservation. The standard training scheme, however, overlooks the potential challenges posed by this coupling and does not explicitly address the associated optimization difficulties. In standard approach, the loss terms corresponding to the AC and CH equations are combined into a single composite loss function, treating the coupled equations in a holistic yet undifferentiated manner. Specifically, Eq.~\eqref{eq:weighted-loss} can be expressed as:
\begin{equation}
    \mathcal{L}_\text{total} = w_\text{AC}\mathcal{L}_\text{AC} + w_\text{CH}\mathcal{L}_\text{CH} +
    w_\text{BC}\mathcal{L}_\text{BC} + w_\text{IC}\mathcal{L}_\text{IC}, \label{eq:weighted-loss-standard}
\end{equation}%
where the subscripts AC, CH, BC, and IC represent the loss contributions from the Allen-Cahn equation, Cahn-Hilliard equation, boundary conditions, and initial conditions, respectively.

\begin{figure}[htbp]
    \subfloat[\label{fig:cosine-similarity-standard-training}]{\includegraphics[width=0.47\textwidth]{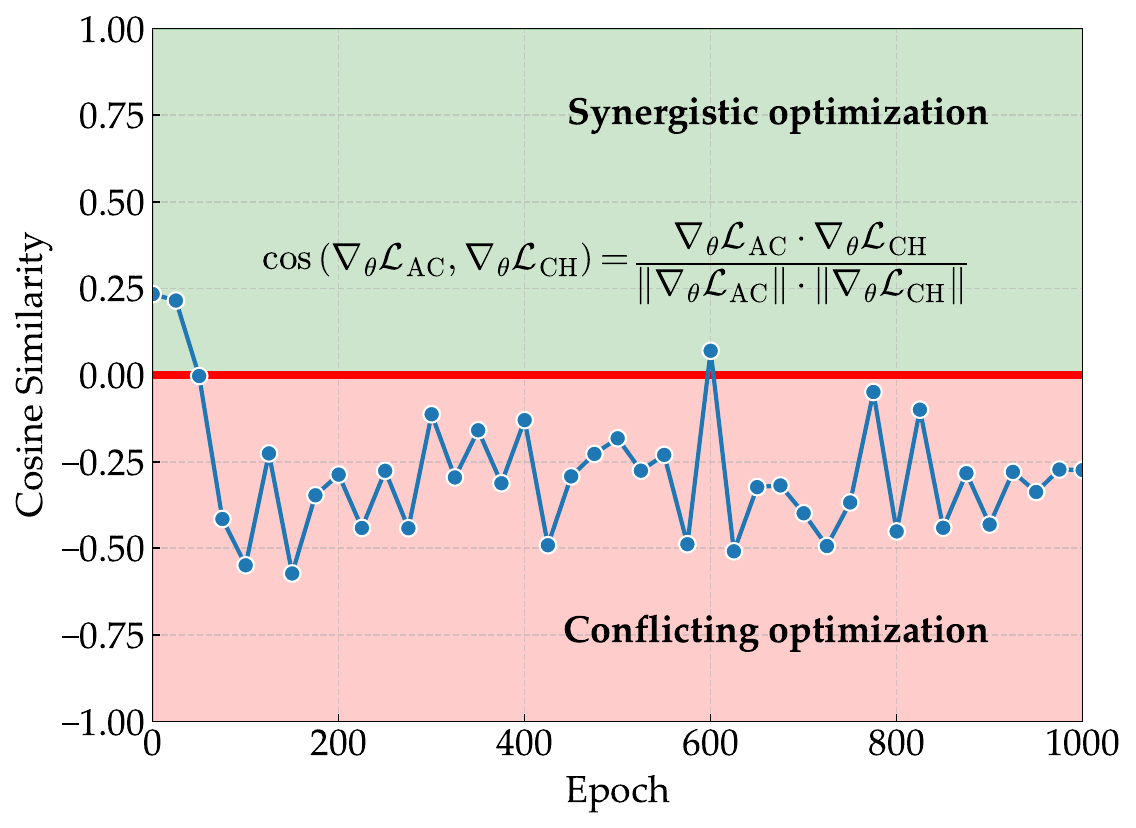}}
    \hfill
    \subfloat[\label{fig:cosine-similarity-stagger-training}]{\includegraphics[width=0.47\textwidth]{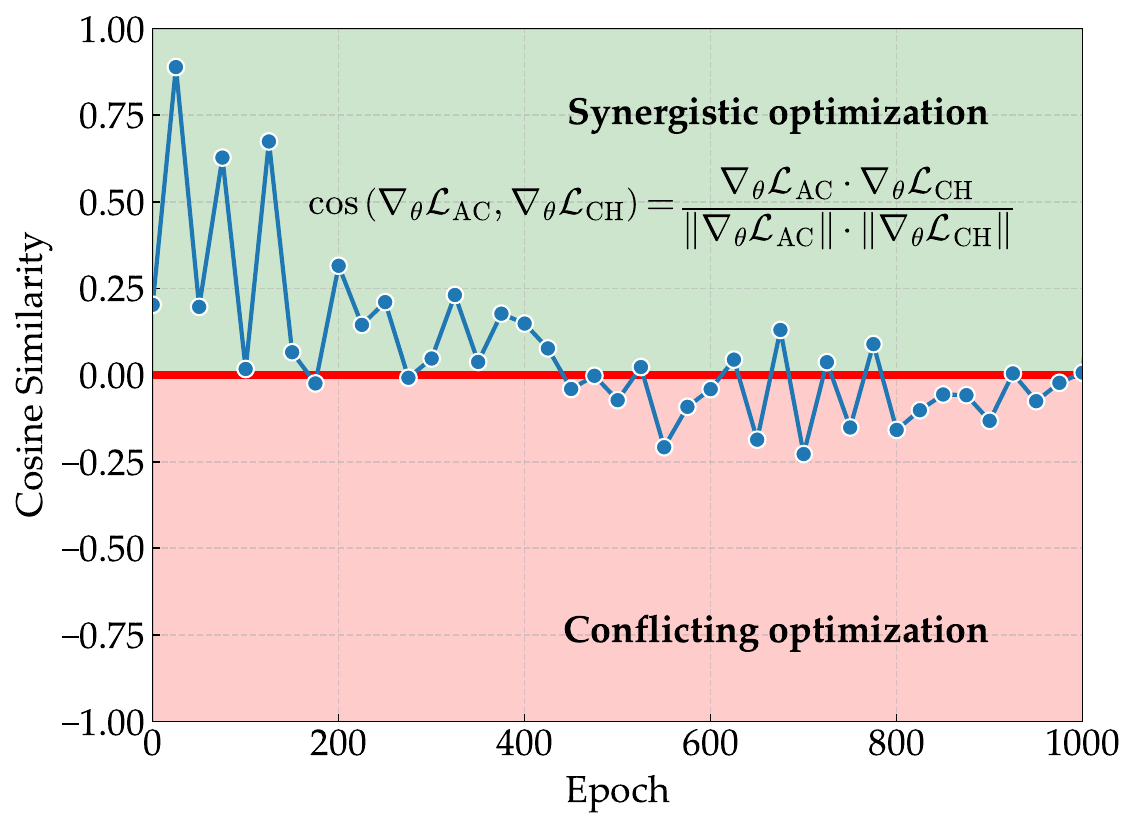}}
    \caption{Cosine similarity between the gradients of the loss terms corresponding to the Allen-Cahn and Cahn-Hilliard equations during (a) standard training scheme and (b) staggered training scheme for the phase field corrosion model.}
\end{figure}

However, this standard scheme fails to account for the distinct characteristics and numerical requirements of the two equations. In multi-task learning scenarios, the simultaneous optimization of the AC and CH equations often leads to gradient conflicts \cite{yuGradientSurgeryMultiTask2020}. This phenomenon is depicted in Figure~\ref{fig:cosine-similarity-standard-training}, where the cosine similarity between the gradients of the AC and CH equations becomes predominantly negative after the initial few epochs. Such negative similarity indicates severe competition between the two equations during optimization, leading to conflicting gradient directions. Consequently, this competitive optimization process would impede convergence and increase the risk of the network being trapped in local minima. For a comprehensive analysis of this phenomenon, we refer readers to Wang {\em et al.} \cite{wangGradientAlignmentPhysicsinformed2025a}.


To overcome the challenges associated with optimizing the coupled AC and CH equations, we propose a novel staggered training scheme that effectively decouples their optimization processes. This approach allows the network to focus on capturing the distinct characteristics of each equation separately, while the coupling between them is implicitly preserved. As illustrated in Figure~\ref{fig:proposed-pinn} a), the training process is divided into two stages, alternating between the AC and CH equations.

In the first stage, the optimization is guided by the Allen-Cahn equation, alongside the boundary and initial conditions, while temporarily disregarding the Cahn-Hilliard equation. The corresponding loss function is formulated as:
\begin{equation}
    \mathcal{L}_\text{stage1} = w_\text{AC}\mathcal{L}_\text{AC} + w_\text{BC}\mathcal{L}_\text{BC} + w_\text{IC}\mathcal{L}_\text{IC}. \label{eq:loss-stage1}
\end{equation}%
In the second stage, the focus shifts to optimizing the Cahn-Hilliard equation, again alongside the boundary and initial conditions, while the Allen-Cahn equation is temporarily excluded. The corresponding loss function is given by:
\begin{equation}
    \mathcal{L}_\text{stage2} = w_\text{CH}\mathcal{L}_\text{CH} + w_\text{BC}\mathcal{L}_\text{BC} + w_\text{IC}\mathcal{L}_\text{IC}. \label{eq:loss-stage2}
\end{equation}%
The training alternates between these two stages, with the optimization of the AC and CH equations staggering every $S_s$ epochs. Here, $S_s$ is a predefined hyperparameter that determines the staggering period between the two stages. The detailed implementation of the staggered training scheme is summarized in Algorithm~\ref{alg:staggered-training}.

\begin{algorithm}[ht]
    \caption{Implementation of the staggered training scheme}
    \label{alg:staggered-training}
    Setup the training process\;
    \While{$s < s_{\text{max}}$}{
        \If{$s \mod (2S_s) = 0$}{
            Resampling the collocation points $\{ \bm x_g^j, t_g^j \}_{j=1}^{N_g}$, $\{ \bm x_b^j, t_b^j \}_{j=1}^{N_b}$, and $\{ \bm x_i^j \}_{j=1}^{N_i}$\;
        }
        \eIf{$s \mod (2S_s) < S_s$}{
            Switch to Stage 1: optimize the Allen-Cahn equation, Let $\text{PDE} := \text{AC}$\;
        }{
            Switch to Stage 2: optimize the Cahn-Hilliard equation, Let $\text{PDE} := \text{CH}$\;
        }
        Forward pass through the network $\mathcal{N}$, compute loss terms $\mathcal{L}_\text{PDE}$, $\mathcal{L}_\text{BC}$, and $\mathcal{L}_\text{IC}$ and determine the weights $w_\text{PDE}$, $w_\text{BC}$, and $w_\text{IC}$\;
        Assemble the loss function:
        \begin{equation*}
            \mathcal{L}_\text{total} = w_\text{PDE}\mathcal{L}_\text{PDE} + w_\text{BC}\mathcal{L}_\text{BC} + w_\text{IC}\mathcal{L}_\text{IC}
        \end{equation*}\\
        Backward pass through the neural network\;
        Step forward in the optimization process: $s \leftarrow s + 1$\;
    }
\end{algorithm}
Here, $s$ denotes the current step, $s_{\text{max}}$ is the maximum number of steps. The weights $w_\text{PDE}$, $w_\text{BC}$, and $w_\text{IC}$ are determined based on a grad norm scaling strategy \cite{wangUnderstandingMitigatingGradient2021a,wangExpertsGuideTraining2023}, which is detailed in Section~\ref{sec:loss-balancing}. It is worth noting that we resample the collocation points every $2S_s$ epochs such that each round of staggered training starts with a fresh set of collocation points while keep the same in each stage. Additionally, the general collocation points set $\{ \bm x_g^j, t_g^j \}_{j=1}^{N_g}$ is constructed as the Cartesian product of collocation points randomly sampled from each dimension.

This staggered approach allows us to optimize each governing equation independently, thereby simplifying the loss function and reducing the complexity of the optimization process. The typical cosine similarity between the gradients of the AC and CH equations during the staggered training scheme is shown in Figure~\ref{fig:cosine-similarity-stagger-training}. Compared with the standard training results, the cosine similarity values during staggered training remain positive in the early stages, indicating a synergistic optimization of the two equations. As training progresses, the cosine similarity gradually decreases and oscillates around zero, reflecting a transition to relatively independent optimization of the equations. This behaviour aligns with the intended design of the staggered training scheme, which mitigates the competitive optimization challenges that occur when both equations are optimized simultaneously.

Here we remark that while our study focuses on the staggered training of AC and CH equations for corrosion problems, the proposed training scheme can be naturally extended to other important coupled PDEs and multi-physics systems, such as reaction-diffusion equations, Navier-Stokes equations, and beyond.


\subsection{Neural network architecture with hard constraints}
\label{sec:neural-network-architecture}

The neural network architecture $\mathcal{N}$ plays a crucial role in determining the performance of PINNs. In this work, the input layer, backbone network, and output layer are carefully designed to enhance the network's capability of approximating the solution field accurately and efficiently. Specifically, the spatial-temporal coordinates $\bm x$ and $t$ are successively fed into the network through a random Fourier feature embedding $\mathcal{F}$ \cite{tancikFourierFeaturesLet2020}, followed by a modified MLP $\mathcal{M}$ \cite{wangUnderstandingMitigatingGradient2021a}, and finally, the output layer with hard constraints $\mathcal{H}$. The overall architecture of the neural network $\mathcal{N}$ can be expressed as:
\begin{equation}
    \mathcal{N} = \mathcal{H} \circ \mathcal{M} \circ \mathcal{F},
    \label{eq:nn-architecture}
\end{equation}%

Since the Fourier feature embedding $\mathcal{F}$ and the modified MLP $\mathcal{M}$ have been extensively discussed in previous studies \cite{tancikFourierFeaturesLet2020,wangEigenvectorBiasFourier2021,wangUnderstandingMitigatingGradient2021}, this section focuses on the hard constraints $\mathcal{H}$ applied in the output layer. The specific implementation details of the Fourier feature embedding and the modified MLP tailored to this problem are presented in \ref{sec:random-fourier-feature-embedding} and \ref{sec:modified-mlp-architecture}, respectively.

We introduce a hard constraint $\mathcal{H}$ that encodes the underlying physics of the phase field corrosion model directly into the network's output layer. The core idea of this encoding  is to explicitly represent the concentration field $c$ according to the KKS corrosion model \cite{kimPhasefieldModelBinary1999}.

Specifically, the solution of the phase field corrosion model involves two field variables: the phase field (order parameter) $\phi$ and the concentration $c$. In the KKS model (see Section~\ref{sec:phase field-model-corrosion} and the references therein), the concentration $c$ is formulated as a mixture of the solid-phase concentration $c_\text{S}$ and the liquid-phase concentration $c_\text{L}$. The phase field $\phi$ and the interpolation function $h(\cdot)$ serve as mixing factors that determine the resulting concentration distribution (Eq.~\eqref{eq:concentration-mixture}).

Rather than directly predicting the phase field $\phi$ and the concentration $c$, we design the network's output to represent the phase field $\phi$ and the liquid-phase concentration $c_\text{L}$. The solid-phase concentration $c_\text{S}$ is then calculated using Eq.~\eqref{eq:equilibrium-concentration-relationship}, and subsequently, the mixture concentration $c$ can be derived from $c_\text{L}$, $c_\text{S}$, and $\phi$ based on Eq.~\eqref{eq:concentration-mixture}, i.e.:
\begin{gather*}
    h(\phi) = -2\phi^3 + 3\phi^2, \\
    c = h(\phi) c_\text{S} + \left[1 - h(\phi)\right] c_\text{L}.
\end{gather*}%
Consequently, this explicit embedding of concentration $c$ in the output layer enhances the physical interpretability of the predictions and thus reduces the optimization difficulties.

Additionally, to further enforce physical constraints, we incorporate bounded activation functions in the network's output layer. For the normalized phase field $\phi$, we restrict its range to $[0,1]$ using the hyperbolic tangent function:
\begin{equation}
    \phi = \frac{1}{2} \tanh(\hat\phi) + \frac{1}{2}.
\end{equation}%
Similarly, we constrain $c_\text{L}$ between $0$ and $1 - c_\text{Se} + c_\text{Le}$ in accordance with Eq.~\eqref{eq:liquid-concentration}, ensuring the mixture concentration $c$ remains within $[0,1]$:
\begin{equation}
    c_\text{L} = \left(1 - c_\text{Se} + c_\text{Le}\right) \cdot \frac{1}{2}\left[ \tanh(\hat{c}_\text{L}) + 1\right].
\end{equation}%
By applying these scaling operators, the network's outputs are guaranteed to fall within a physically meaningful range.

\begin{algorithm}[H]
    \caption{Implementation of the hard constraints $\mathcal{H}$ in the output layer}
    \label{alg:hard-constraints}
        \textbf{Input:} Spatial-temporal coordinates $\bm x, t$\;
        \textbf{Output:} Phase field $\phi$, concentration $c$\;
        \textbf{Initialize:} Layers of input embeddings $\mathcal{F}$ and backbone network $\mathcal{M}$\;
        Perform a forward pass through the network and apply normalization to obtain $\phi$ and $\hat c_\text{L}$:
        \begin{equation*}
            \hat c_\text{L}, \phi = \frac{1}{2}\tanh\left\{\mathcal{M}\left[\mathcal{F}\left(\bm x, t\right)\right]\right\} + \frac{1}{2}
        \end{equation*}\\
        Re-scale the liquid-phase concentration $\hat c_\text{L}$ to the range $[0, 1-c_\text{Se}+c_\text{Le}]$:
        \begin{equation*}
            c_\text{L} = \left(1 - c_\text{Se} + c_\text{Le}\right) \hat c_\text{L}
        \end{equation*}\\
        Derive the solid-phase concentration $c_\text{S}$ based on the equal chemical potential assumption:
        \begin{equation*}
            c_\text{S} = c + \left(c_\text{Se} - c_\text{Le}\right)
        \end{equation*}\\
        Compute the concentration $c$ according to the mixture assumption using $c_\text{S}$, $c_\text{L}$, and $\phi$:
        \begin{equation*}
            c = h(\phi) c_\text{S} + \left[1 - h(\phi)\right] c_\text{L}
        \end{equation*}\\
        Return the combined output $\phi$ and $c$.
\end{algorithm}

A schematic and detailed implementation of the hard constraints in the output layer are presented in Figure~\ref{fig:proposed-pinn} b) and Algorithm~\ref{alg:hard-constraints}, respectively.

\subsection{Training pipeline}

Combining all the components described above, the training pipeline of the proposed Sharp-PINN framework is summarized in Algorithm~\ref{alg:training-process}. The neural network $\mathcal{N}$ is defined as the composition of the random Fourier feature embedding $\mathcal{F}$, the modified MLP architecture $\mathcal{M}$, and the hard constraint output layer $\mathcal{H}$. The loss function is assembled based on the staggered training scheme, with the PDE term switching between the AC and CH equations. The weights of the loss terms are determined using the grad norm weighting scheme to balance the interplay of individual losses.
The implementation details of applying the Sharp-PINN framework to the phase field corrosion model are discussed in \ref{sec:implementation-details}. In addition to the standard hyperparameters commonly used in neural networks, other parameters specific to the Sharp-PINN framework are summarized in Table~\ref{tab:hyperparameters} unless otherwise stated.

\begin{algorithm}
    \caption{Trianing pipeline of the Sharp-PINNs framework}
    \label{alg:training-process}
    Initialize the neural network $\mathcal{N}$ with the random Fourier feature embedding $\mathcal{F}$ (\ref{sec:random-fourier-feature-embedding}), the modified MLP architecture $\mathcal{M}$ (\ref{sec:modified-mlp-architecture}), and the hard constraint output layer $\mathcal{H}$ (Section~\ref{sec:neural-network-architecture}):
    \begin{equation*}
        \mathcal{N} = \mathcal{H} \circ \mathcal{M} \circ \mathcal{F}
    \end{equation*}\\
    Define the formulation of non-dimensionalized PDE residuals using Eq.~\eqref{eq:governing-equations-non-dimensional-parameters}, initial condition residuals using Eq.~\eqref{eq:phixd} and Eq.~\eqref{eq:cdiffuse}, and boundary condition residuals\;
    \While{$s < s_{\text{max}}$}{
        \If {$s \mod (2S_s) = 0$}{
            Resample the collocation points $\{ \bm x_g^j, t_g^j \}_{j=1}^{N_g}$, $\{ \bm x_b^j, t_b^j \}_{j=1}^{N_b}$, and $\{ \bm x_i^j \}_{j=1}^{N_i}$\;
        }
        \If {$s \mod S_s = 0$}{
            Switch the staggered training stage (\ref{sec:staggered-training}): $\text{PDE} := \text{AC}$ or $\text{CH}$\;
        }
        \textbf{Forward propagation:} Pass the collocation points through the network $\mathcal{N}$ and compute the loss terms $\mathcal{L}_\text{PDE}$, $\mathcal{L}_\text{BC}$, and $\mathcal{L}_\text{IC}$\;
        Update the weights $w_\text{PDE}$, $w_\text{BC}$, and $w_\text{IC}$ using the grad norm weighting scheme (\ref{sec:loss-balancing})\;
        Assemble the loss function using Eq.~\eqref{eq:loss-stage1} or Eq.~\eqref{eq:loss-stage2}:
        \begin{equation*}
            \mathcal{L}_\text{total} = w_\text{PDE}\mathcal{L}_\text{PDE} + w_\text{BC}\mathcal{L}_\text{BC} + w_\text{IC}\mathcal{L}_\text{IC}
        \end{equation*}\\
        \textbf{Backward propagation:} Compute the gradients of the loss function and update the network weights\;
        Step forward in the optimization process: $s \leftarrow s + 1$\;
    }
\end{algorithm}%

\section{Results and discussion}
\label{sec:results-and-discussion}



In this section, we extensively validate the capability of the proposed Sharp-PINN framework by a series of benchmark cases of phase field corrosion problems. The benchmark cases include two-dimensional corrosion problems with two and three semi-circular pits, as well as three-dimensional corrosion problems with one and two pits. We also perform comprehensive ablation studies to investigate the effectiveness and necessity of each component of the Sharp-PINN framework on two- and three-dimensional corrosion problems with two initial pits. 

All codes associated with this Section are developed using the \texttt{Python} programming language and the \texttt{PyTorch} deep learning library. For each case study, a reference solution is obtained via the finite element method implemented using the \texttt{Python} package \texttt{FEniCS} to ensure reliable benchmarks for comparison. All the benchmark cases and the corresponding reference solutions are executed on a computing server with an AMD EPYC 7543 CPU with 32 cores and 80 GB of memory and an NVIDIA A40 GPU with 48 GB of memory. We also provide a detailed description of the FEM implementation in \ref{sec:fem-implementation}. The hyperparameters for training our Sharp-PINN models are summarized in Table~\ref{tab:hyperparameters}, and the physical parameters for the phase field corrosion model are listed in Table~\ref{tab:parameters}.



\begin{table}[H]
    \caption{State-of-the-art benchmark results for the Sharp-PINNs framework.}
    \label{tab:benchmark-results}
    \begin{tblr}{
        width=\textwidth,
        colspec={Q[c,1.5]Q[c,1.5]Q[c,1]Q[c,1]Q[c,1]},
        rowspec={Q[m]Q[m]Q[m]Q[m]Q[m]},
        row{1}={font=\bfseries}}
        \hline
        {Benchmark\\case} & {Representative\\result} & {Absolute\\$L^2$ error} & {FEM\\time ($\mathrm{min}$)} & {PINN\\time ($\mathrm{min}$)} \\
        \hline
        \ref{sec:2d-2pits} 2D 2-pits& \includegraphics[height=3em,align=c]{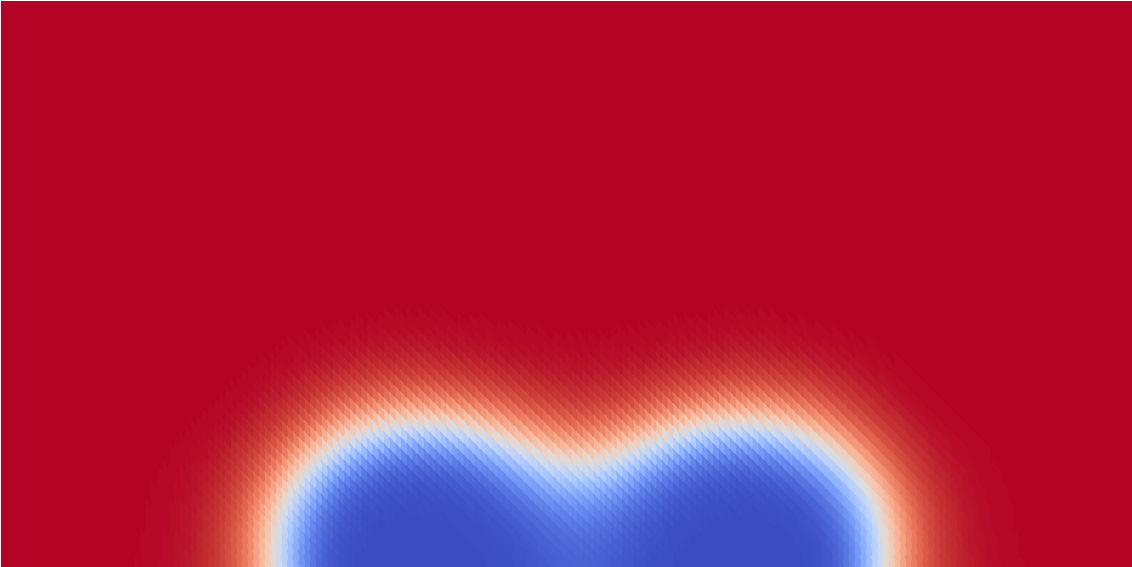} &$6.066\times10^{-4}$ & $1.62$ & $7.68$ \\
        \ref{sec:2d-3pits} 2D 3-pits& \includegraphics[height=3em,align=c]{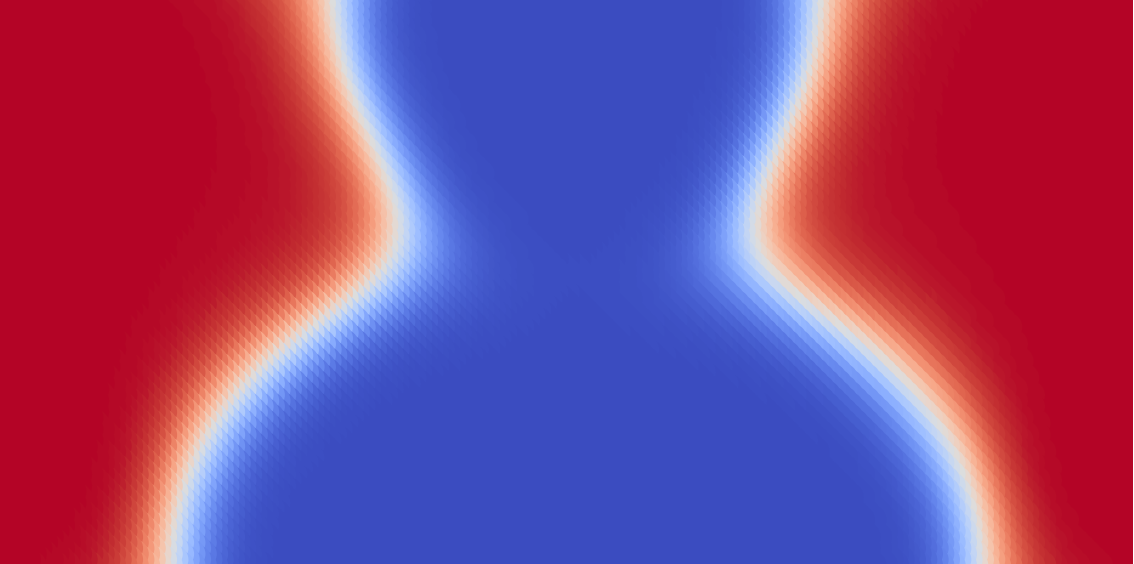} & $1.755\times10^{-3}$ & $1.85$ & $8.50$ \\
        \ref{sec:3d-1pit} 3D 1-pit& \includegraphics[height=3em,align=c]{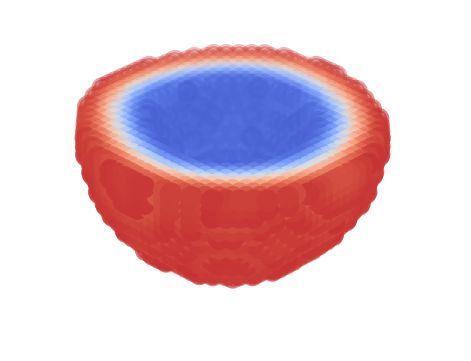}&$1.556\times10^{-3}$ & $107.70$ & $17.05$ \\
        \ref{sec:3d-2pits} 3D 2-pits&\includegraphics[height=3em,align=c]{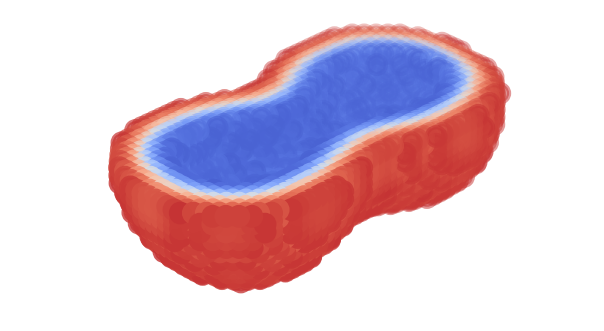}& $2.834\times10^{-3}$ & $197.23$  & $18.40$ \\
        \hline
    \end{tblr}
\end{table}

Table~\ref{tab:benchmark-results} summarizes the benchmark results achieved by the Sharp-PINNs framework, highlighting its performance in terms of absolute $L^2$ error and computational time comparisons with the finite element method. The reported $L^2$ errors consistently confirm the accuracy of the PINN solutions across all cases. Notably, for the three-dimensional problems, the Sharp-PINN models achieve a substantial reduction in computational time compared to the finite element method, underscoring the efficiency of the Sharp-PINNs and their potential in engineering applications. Detailed results and discussion for each benchmark test are provided in the following subsections.



\subsection{Two-dimensional corrosion with two-pits interactions}
\label{sec:2d-2pits}

\begin{figure}[htbp]
    \centering
    \includegraphics[width=0.8\textwidth]{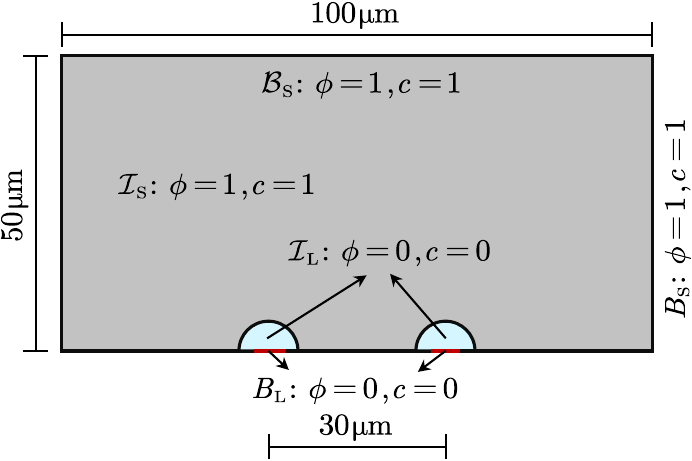}
    \caption{Two-dimensional corrosion with two-pits interactions: Geometric setup, initial conditions, and boundary conditions.}
    \label{fig:icbc-2d2pits}
\end{figure}

For the first case study, we consider a two-dimensional corrosion problem with two semi-circular pits. The initial and boundary conditions for the phase field $\phi$ and concentration $c$ are depicted in Figure~\ref{fig:icbc-2d2pits}. The spatial domain is defined as $\mathcal{D} = [-50, 50] \times [0, 50]\,\mathrm{\mu m}^2$, with the temporal domain spanning $\mathcal{T}=[0, 10]\,\mathrm{s}$. Initially, two semi-circular pits, each with a radius of $r = 5\,\mathrm{\mu m}$, are positioned along the bottom boundary of the domain, separated by a center-to-center distance of $30\,\mathrm{\mu m}$. The initial and boundary conditions specify $\phi=0$ and $c=0$ for the liquid phase, while $\phi = 1$ and $c = 1$ are assigned for the solid phase.

We shall first develop a baseline model and then discuss the impact of each component of the proposed Sharp-PINN framework through a series of ablation studies. The baseline model integrates all the components of the Sharp-PINNs framework, including the staggered training scheme, hard constraints, modified MLP, and Fourier features. The hyperparameters are set to the default values listed in Table~\ref{tab:hyperparameters}. 

\begin{figure}[ht]
    \centering
    \includegraphics[width=\textwidth]{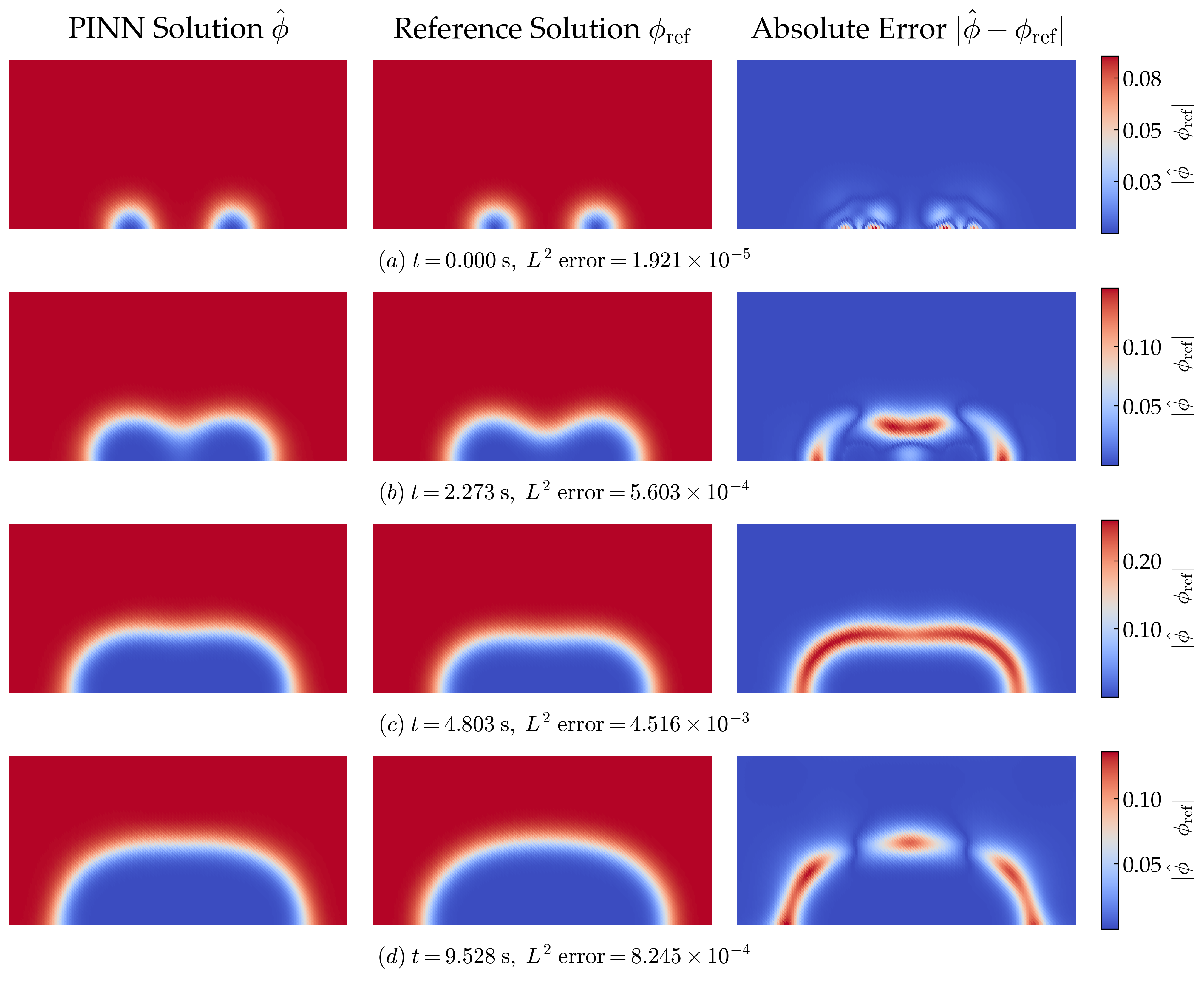}
    \caption{Two-dimensional corrosion with two-pits interactions. Contours of field variable $\phi$ obtained from PINNs ($\hat \phi$), \texttt{FEniCS} ($\phi_\text{ref}$), and their absolute error ($|\hat \phi - \phi_\text{ref}|$).}
    \label{fig:2d-2pits-phi-fields}
\end{figure}

We train the Sharp-PINN model for $1,000$ steps using the Adam optimizer with an initial learning rate of $5.0\times10^{-4}$, decaying exponentially by $0.9$ every $100$ epochs. 
Figure~\ref{fig:2d-2pits-phi-fields} compares the Sharp-PINNs solution $\hat \phi$, the \texttt{FEniCS} reference solution $\phi_\text{ref}$, and their absolute difference $|\hat \phi - \phi_\text{ref}|$ at four representative times.
The results demonstrate that Sharp-PINNs can accurately capture the complex evolution of the phase variable. The maximum error occurs within the localized interface regions, where the field variable undergoes rapid changes, while the error in the bulk regions remains relatively low. The maximum $L^2$ error of the phase field $\phi$ is $4.516\times10^{-3}$ at $t=4.803\,\mathrm{s}$, indicating an excellent agreement between the Sharp-PINNs solution and the reference solution. Remarkably, the Sharp-PINN framework accurately captures the pits interaction and coalescence process, a task that has been challenging for previous PINN-based approaches \cite{chenPFPINNsPhysicsinformedNeural2025}.

\begin{figure}[htbp]
    \centering
    \includegraphics[width=0.6\textwidth]{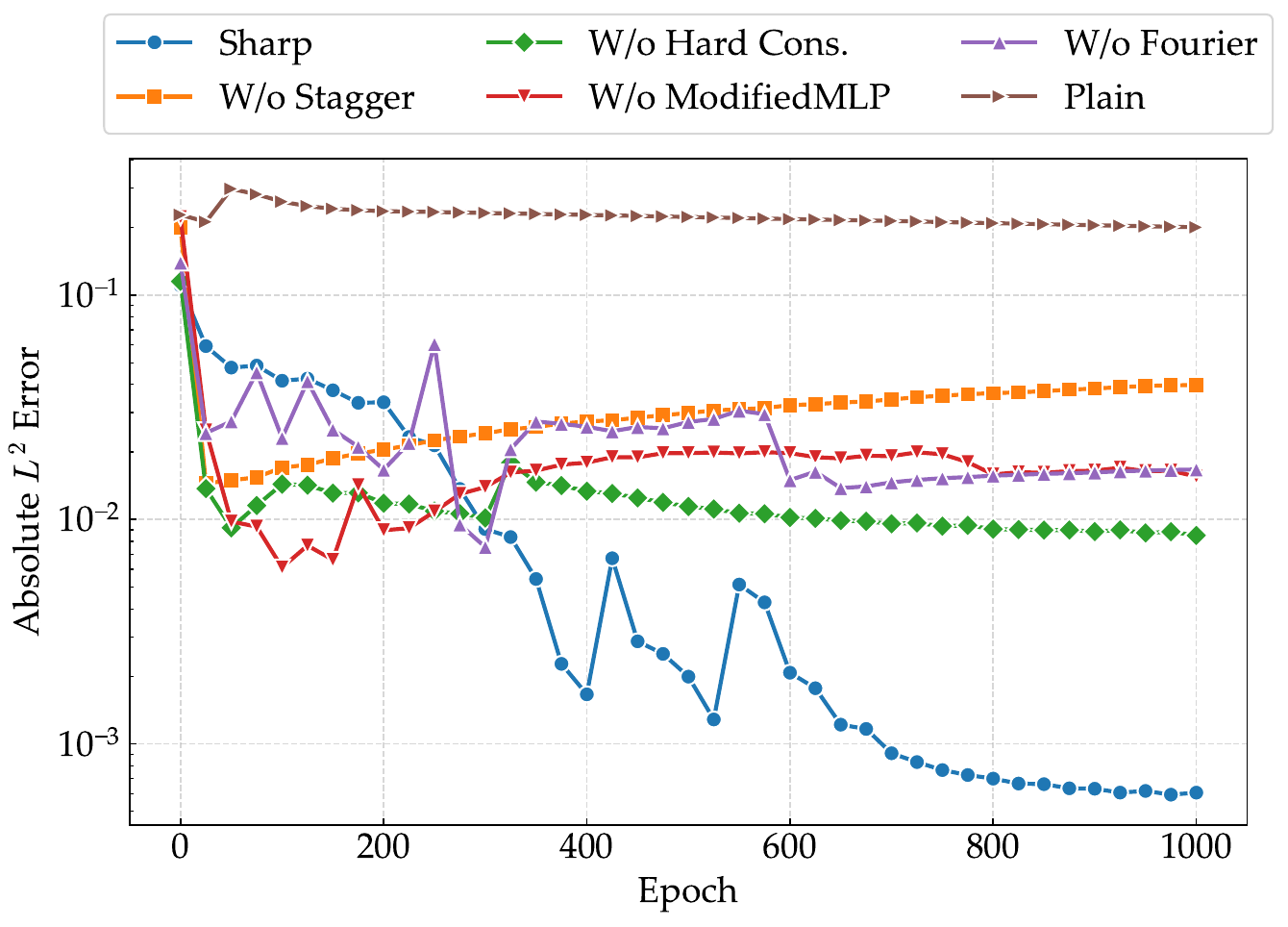}
    \caption{Two-dimensional corrosion with two-pits interactions. Convergence of absolute $L^2$ error of ablation studies. The abbreviations in the legend represent PINN models with different ablation configurations. Sharp: the proposed baseline model with all components enabled. W/o Stagger: PINN model without the staggered training scheme. W/o Hard Cons.: PINN model without hard constraints. W/o ModifiedMLP: PINN model with the standard MLP architecture. W/o Fourier: PINN model without Fourier feature embedding. Plain: the standard PINN model without any modifications.}
    \label{fig:2d-2pits-l2-error-history}
\end{figure}

To demonstrate the effectiveness and necessity of each component in the Sharp-PINNs framework, we perform a series of ablation studies by systematically disabling individual components one at a time while keeping the remaining components active and the hyperparameters unchanged. The ablation configurations and corresponding performance metrics are summarized in Table~\ref{tab:ablation-studies-2d2pits}. The convergence history of the absolute $L^2$ error for each ablation case is depicted in Figure~\ref{fig:2d-2pits-l2-error-history}. The results clearly show that our Sharp-PINN model, with all components enabled, achieves the lowest average $L^2$ error of $6.066 \times 10^{-4}$.  In contrast, disabling any component results in a substantial increase in error (exceeding one order of magnitude). It implies the critical role of each component plays in enhancing the model's performance.  It is worth noting that the most significant performance degradation occurs when the staggered training scheme is disabled, leading to an absolute $L^2$ error of $3.974 \times 10^{-2}$ and an increase in training time of approximately $1.5$ minutes. This highlights the benefits of the staggered training scheme in improving both accuracy and efficiency. 
Among all components, the modified MLP architecture is the most computationally demanding, as disabling it reduces the training time by about one-third. However, this trade-off is well justified by the significant performance gains it delivers, demonstrating the value of the additional computational cost.

\begin{table}[htbp]
    \centering
    \caption{Two-dimensional corrosion with two-pits interactions. Ablation configurations and performance metrics. The abbreviations in the first column are consistent with those in Figure~\ref{fig:2d-2pits-l2-error-history}.}
    \label{tab:ablation-studies-2d2pits}
    \resizebox{\textwidth}{!}{
    \begin{tblr}{
        colspec={Q[c]Q[c]Q[c]Q[c]Q[c]Q[c]Q[c]Q[c]Q[c]},
        row{1-2}={font=\bfseries},}
        \hline
        \SetCell[r=2,c=1]{m} Abbreviations && \SetCell[r=1,c=4]{c} Ablation Settings &&&& & \SetCell[r=1,c=2]{c} Performance \\
        \cline{1,3-6,8-9}
        &&{Stagger\\Training} & {Hard\\Constraints} & {Modified\\MLP} & {Fourier\\Features} && {Absolute\\$L^2$ error} & {Run time\\($\text{min}$)}  \\
        \hline
        Sharp&&\checkmark & \checkmark & \checkmark & \checkmark &&  $6.066\times10^{-4}$  & $7.68$ \\
        W/o Stagger&&\xmark & \checkmark & \checkmark & \checkmark && $3.974\times10^{-2}$ & $9.15$ \\
        W/o Hard Cons.&&\checkmark & \xmark & \checkmark & \checkmark && $8.494\times10^{-3}$ & $7.35$ \\
        W/o ModifiedMLP&&\checkmark & \checkmark & \xmark & \checkmark && $1.554\times10^{-2}$ & $2.88$ \\
        W/o Fourier&&\checkmark & \checkmark & \checkmark & \xmark && $1.671\times10^{-2}$ & $7.42$ \\
        Plain&&\xmark & \xmark & \xmark & \xmark && $2.006\times10^{-1}$ & $3.03$ \\
        \hline
    \end{tblr}
    }
\end{table}

\begin{figure}[H]
    \centering
    \includegraphics[width=0.9\textwidth]{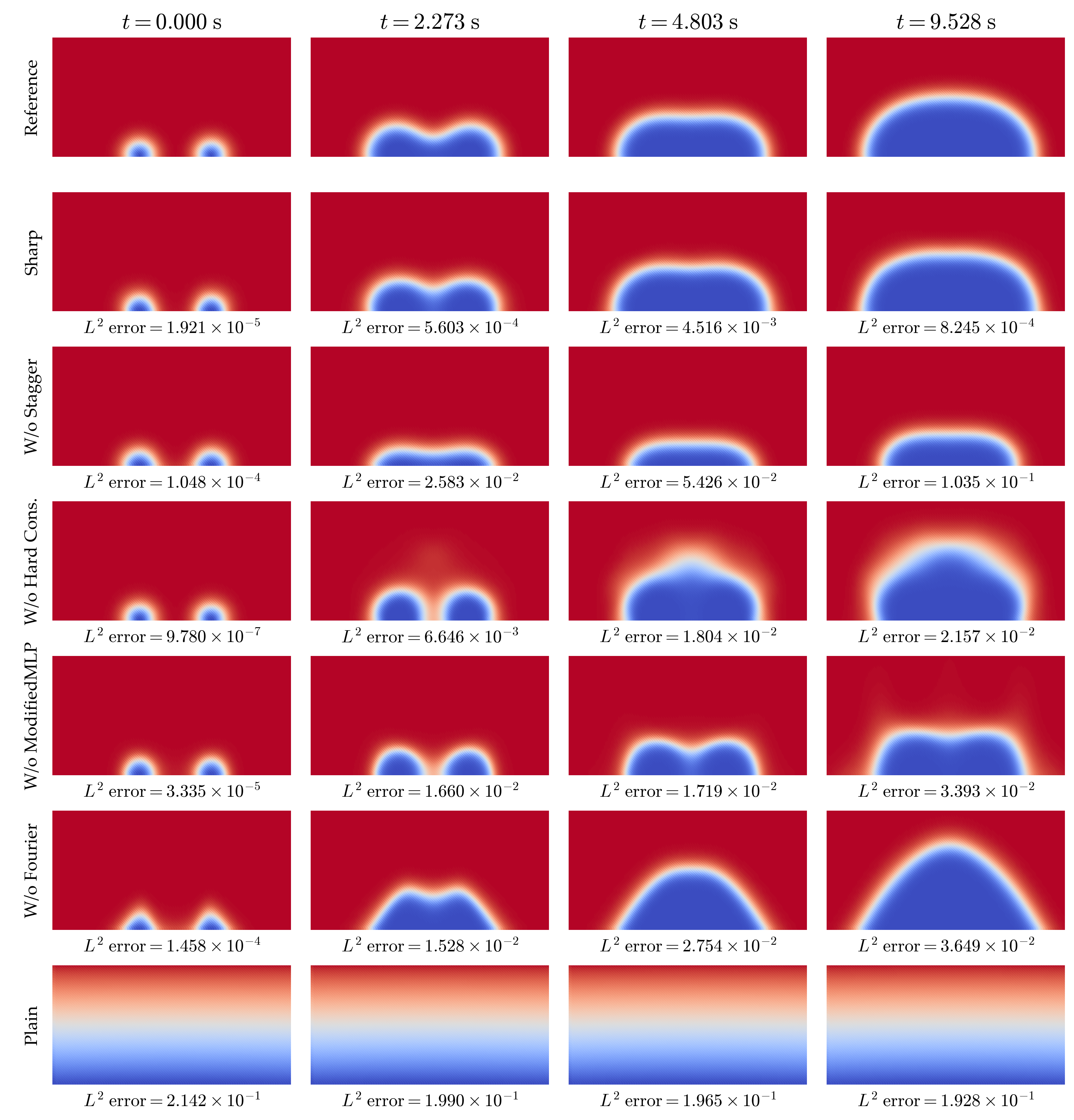}
    \caption{Two-dimensional corrosion with two-pits interactions. Comparison of the predicted phase field variable $\phi$ (pitting growth) obtained from PINNs of different ablation settings. The abbreviations in the left column are consistent with Figure~\ref{fig:2d-2pits-l2-error-history}.}
    \label{fig:2d-2pits-phi-fields-comparison-all}
\end{figure}

A comprehensive comparison of the predicted field variable $\phi$ (pitting growth) obtained  by our ablation studies is illustrated in Figure~\ref{fig:2d-2pits-phi-fields-comparison-all}, providing a visual analysis of the failure modes associated with each ablated model. The main observations are summarized as follows:
\begin{itemize}
    \item The baseline model demonstrates excellent agreement with the reference solution, accurately capturing the evolution of $\phi$. 
    \item In contrast, when the staggered training scheme is disabled, the model still captures the coalescence result but exhibits premature growth of the pits toward each other. Additionally, growth in the direction perpendicular to their connecting line is significantly underestimated, indicating a false convergence direction in the absence of the staggered training scheme. 
    \item When the hard constraints are removed, major errors emerge in the interface regions, particularly after the coalescence. The interface profiles are poorly resolved, highlighting the critical role of hard constraints in capturing sharp interface transitions. 
    \item Similarly, disabling the modified MLP architecture results in the model's inability to accurately capture the coalescence process, leading to inaccurate post-coalescence morphology. This underscores the insufficient representational capacity of the standard MLP architecture. 
    \item Removing the Fourier feature embedding prevents the model from even capturing the initial semi-circular pits' morphology, emphasizing the importance of Fourier features in representing the spatial-temporal patterns of the phase field. 
    \item Finally, the standard PINN model without any modifications entirely fails to capture the phase field evolution, with the predicted $\phi$ deviating significantly from the reference solution.
\end{itemize}%
In conclusion, these ablation studies decisively demonstrate the necessity and effectiveness of each component in the Sharp-PINNs framework.

\subsection{Two-dimensional corrosion with three-pits interactions}
\label{sec:2d-3pits}

\begin{figure}[htbp]
    \centering
    \includegraphics[width=0.7\textwidth]{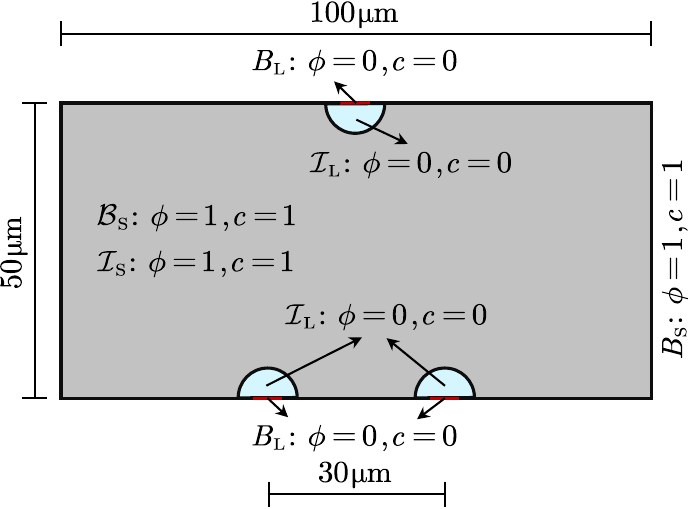}
    \caption{Two-dimensional corrosion with three-pits interactions. Initial condition and boundary conditions for the phase field $\phi$ and concentration $c$.}
    \label{fig:icbc-2d3pits}
\end{figure}

For the second case study, we extend the two-dimensional corrosion problem to the coalescence of three semi-circular pits. Besides the two initial pits on the bottom side, a third pit is placed at the centre of the top side, with the same radius of $r = 5\,\mathrm{\mu m}$. The spatial and temporal domains are identical to the previous case, with the detailed initial and boundary conditions shown in Figure~\ref{fig:icbc-2d3pits}. 

\begin{figure}[htbp]
    \centering
    \includegraphics[width=\textwidth]{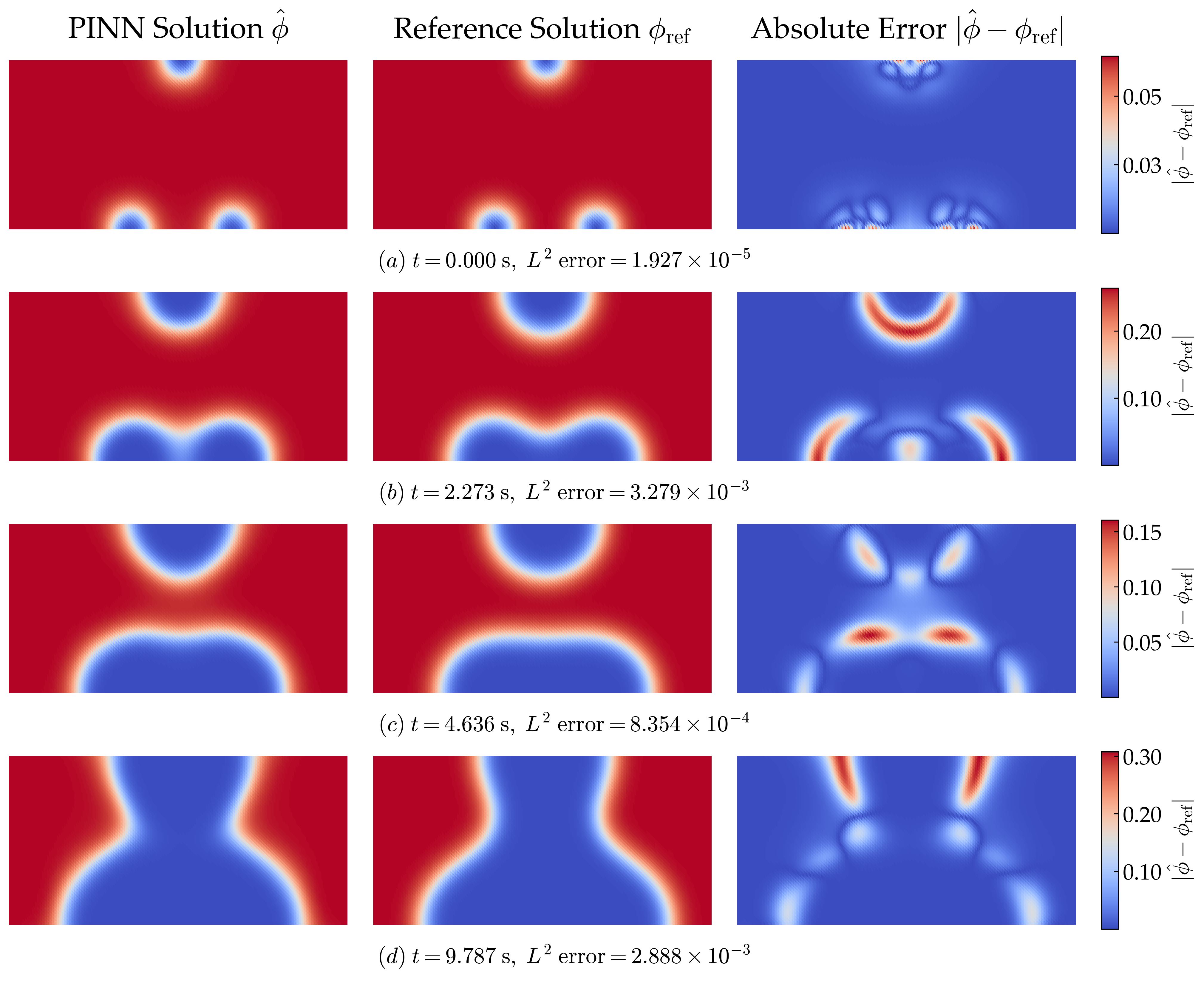}
    \caption{Two-dimensional corrosion with three-pits interactions. Contours of phase field variable $\phi$ obtained from PINNs ($\hat \phi$), \texttt{FEniCS} ($\phi_\text{ref}$), and their absolute error ($|\hat \phi - \phi_\text{ref}|$).}
    \label{fig:2d-3pits-phi-fields}
\end{figure}

We train the model with all components enabled for $1,000$ steps using default hyperparameters listed in Table~\ref{tab:hyperparameters}. Results of the PINN solution $\hat \phi$, the reference solution $\phi_\text{ref}$, and the corresponding absolute error $|\hat \phi - \phi_\text{ref}|$ at four representative time points are shown in Figure~\ref{fig:2d-3pits-phi-fields}. The evolution process involving three-pits coalescence is accurately captured by the Sharp-PINN model. The prediction errors are primarily concentrated at the interface boundaries, with the maximum absolute error of $3.279\times10^{-3}$ observed at $t=2.273\,\mathrm{s}$. These results demonstrate that the Sharp-PINN framework can effectively model complex multi-pit interactions in real-world engineering applications.

\subsection{Three-dimensional prediction of a single pit}
\label{sec:3d-1pit}

We now turn to more challenging scenarios: semi-circular pitting corrosion in three-dimensional space. One case study focuses on the evolution of a single pit, while the other examines interactions between two pits. Conventional finite element methods face significant challenges in solving such complex three-dimensional problems due to high computational costs and potential convergence issues \cite{GeneralizedFiniteElement2000}.

\begin{figure}[htbp]
    \centering
    \includegraphics[width=0.6\textwidth]{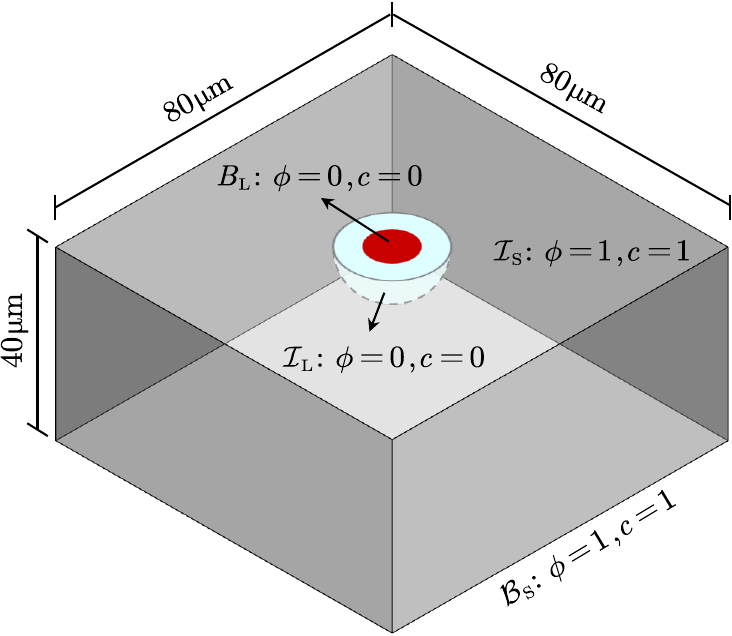}
    \caption{Three-dimensional prediction of a single pit. Initial condition and boundary conditions for the phase field $\phi$ and concentration $c$.}
    \label{fig:icbc-3d1pit}
\end{figure}

For the first three-dimensional case study, we consider a single initial semi-circular pit. The spatial domain is defined as $\mathcal{D} = [-40, 40] \times [-40, 40] \times [0, 40]\,\mathrm{\mu m}^3$, with the temporal domain spanning $\mathcal{T}=[0, 10]\,\mathrm{s}$. The initial pit is positioned on the center of the top side, with a radius of $r = 10\,\mathrm{\mu m}$. The initial and boundary conditions for the phase field $\phi$ and concentration $c$ are depicted in Figure~\ref{fig:icbc-3d1pit}.

Since the spatial dimensionality increases to three, the general sampling points are increased to $N_g = 20\times 20 \times 15\times 30= 180,000$ for $x$, $y$, $z$, and $t$ dimensions, respectively. All other hyperparameters remain the default values listed in Table~\ref{tab:hyperparameters}. We train the model for $1,000$ steps with an initial learning rate of $5.0\times10^{-4}$, which exponentially decays by a factor of $0.9$ for every $100$ epoch. 

\begin{figure}[htbp]
    \centering
    \includegraphics[width=\textwidth]{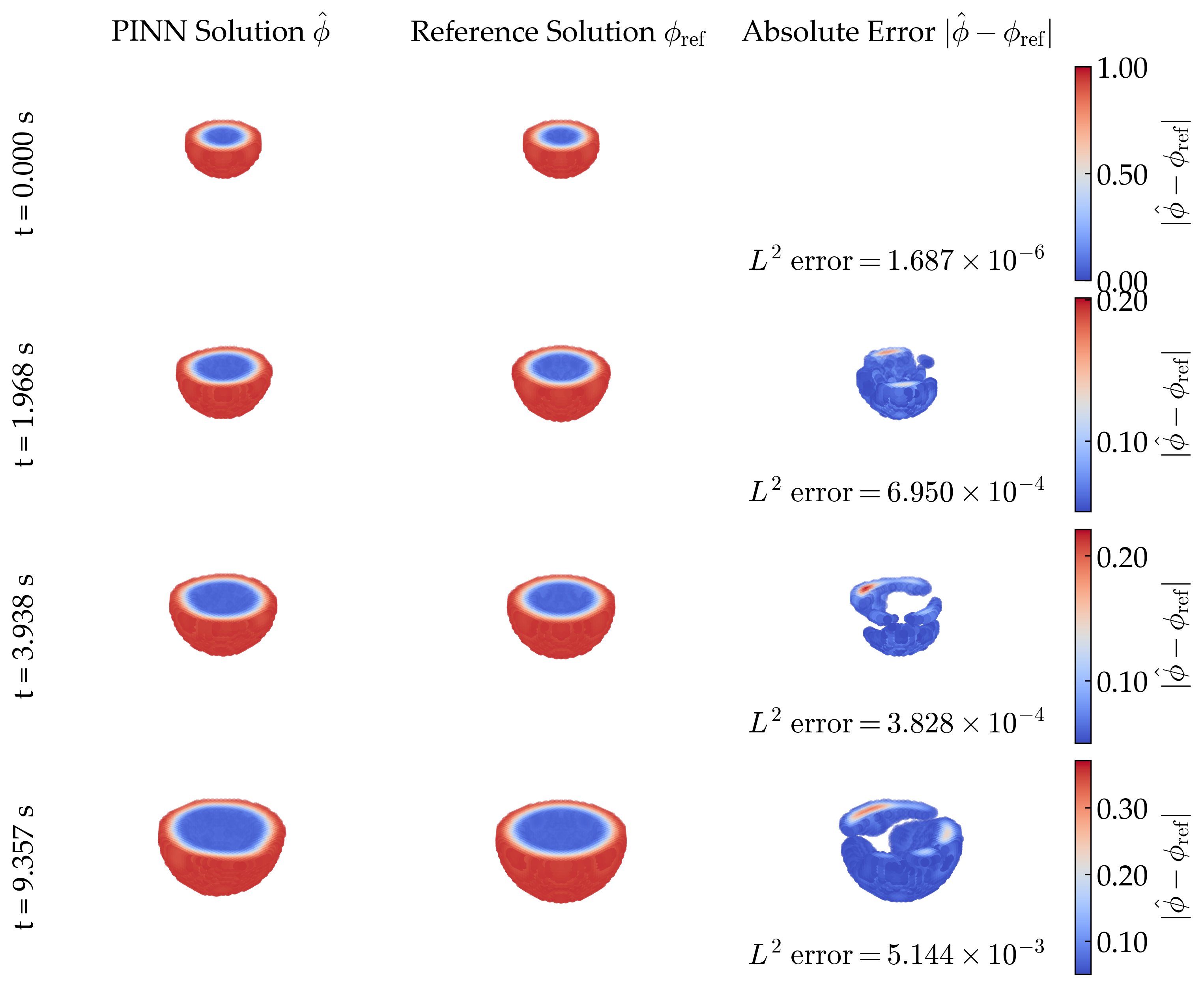}
    \caption{Three-dimensional prediction of a single pit. Contours of phase field variable $\phi$ obtained from PINNs ($\hat \phi$), \texttt{FEniCS} ($\phi_\text{ref}$), and their absolute error ($|\hat \phi - \phi_\text{ref}|$). For visibility, the contours of $\hat \phi$ and $\phi_\text{ref}$ display only the interface region ($0.05 \leq \hat{\phi}, \phi_\text{ref} \leq 0.95$), while the error contours highlight the region where the absolute error exceeds $0.05$.}
    \label{fig:3d-1pits-phi-fields}
\end{figure}

Figure~\ref{fig:3d-1pits-phi-fields} presents a comparative visualization of the PINN-predicted solution $\hat \phi$, the ground truth $\phi_\text{ref}$, and their absolute difference $|\hat \phi - \phi_\text{ref}|$ at four representative time steps.
The model accurately predicts the temporal evolution of the $\phi$ field, achieving strong concordance with the ground truth, with the peak absolute error of $5.144 \times 10^{-3}$ occurring at $t = 9.357\,\mathrm{s}$.
These results validate our PINN framework's effectiveness for three-dimensional phase field corrosion problems. Moreover, our approach demonstrates substantial computational advantages, with a total runtime of $17.05\,\mathrm{min}$ compared to $107.70\,\mathrm{min}$ for the FEM-based baseline - a 6x speedup.

\begin{figure}[htbp]
    \centering
    \resizebox{\textwidth}{!}{
    \subfloat[\label{fig:3d-1pit-stagger-period}]{\includegraphics[height=3cm]{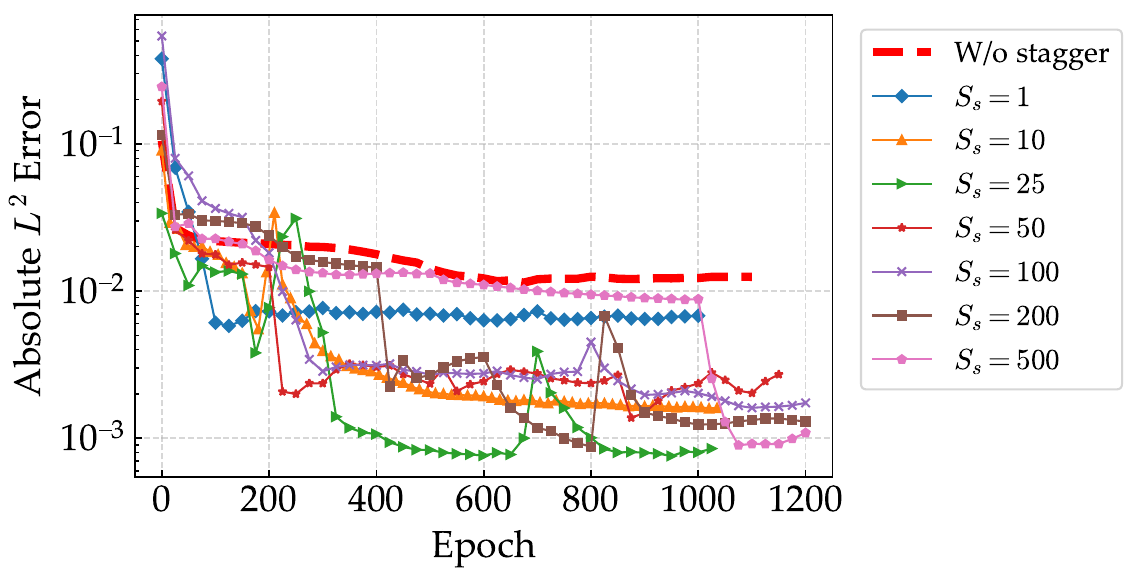}}
    \subfloat[\label{fig:3d-1pit-stagger-l2-with-Ss}]{\includegraphics[height=3cm]{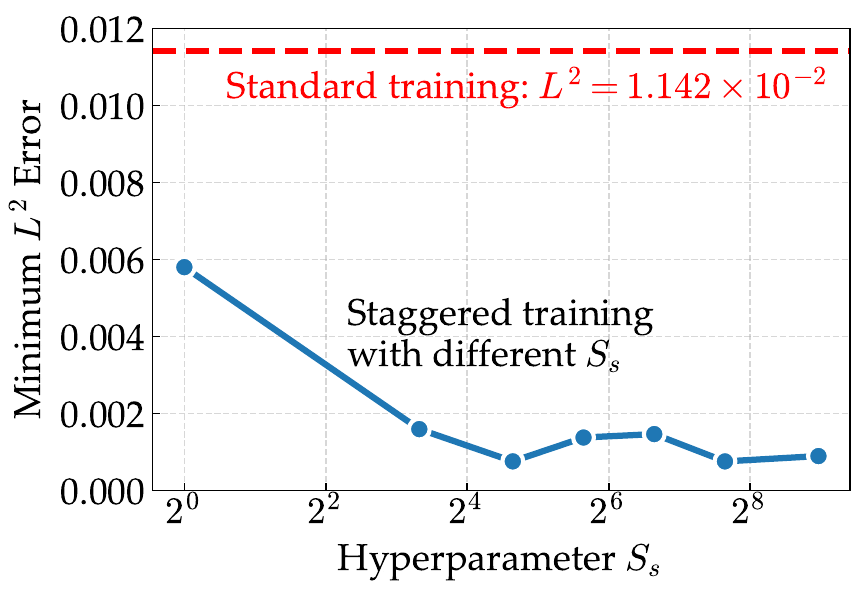}}
    }
    \caption{Three-dimensional prediction of a single pit. (a) Convergence of absolute $L^2$ error during the staggered training scheme with different periods $S_s$. (b) Maximum absolute $L^2$ error with different staggered training periods $S_s$.}
\end{figure}

To investigate the impact of staggered periods $S_s$ on the model's performance, we conduct a series of experiments with different staggered training periods, ranging from $1$ to $500$. The convergence history of the absolute $L^2$ error during the staggered training scheme with varying periods is shown in Figure~\ref{fig:3d-1pit-stagger-period}. A line plot depicting the minimum absolute $L^2$ error achieved with different staggered training periods is presented in Figure~\ref{fig:3d-1pit-stagger-l2-with-Ss}. It can be observed that, except for the period $S_s = 1$, models with staggered training schemes converge to similar minimum errors around $1.0\times 10^{-3}$, approximately one order of magnitude lower than the standard training without staggered schemes. Furthermore, models with $S_s \in [25, 100]$ demonstrate relatively faster and more stable convergence compared to other periods. Another key observation is that the error decreases precisely at the staggering points between the AC and CH equations, especially for models with higher $S_s$, such as epoch $1000$ for $S_s = 500$ and epoch $400$ for $S_s = 200$. This further emphasis the critical role of the staggered training scheme for the coupled AC-CH equations.

\subsection{Three-dimensional corrosion with two-pits interactions}
\label{sec:3d-2pits}

\begin{figure}[ht]
    \centering
    \includegraphics[width=0.8\textwidth]{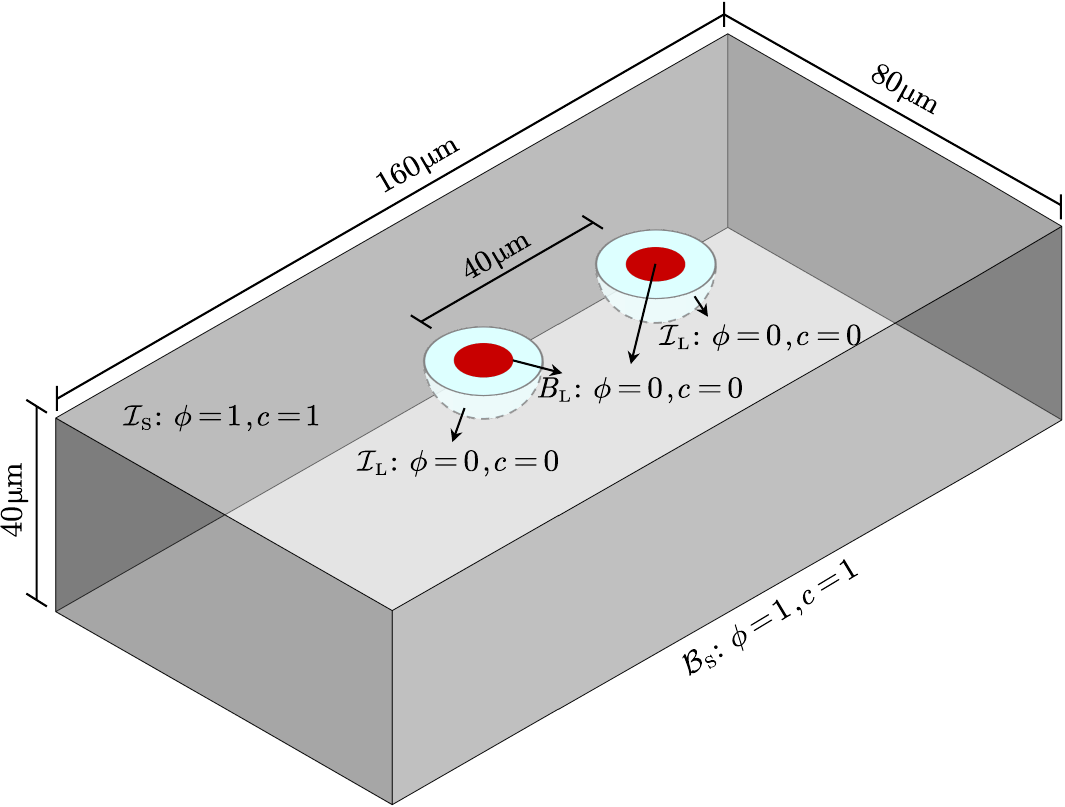}
    \caption{Three-dimensional semi-circular pitting corrosion with two pits interactions. Initial and boundary conditions for the phase field $\phi$ and concentration $c$.}
    \label{fig:icbc-3d2pits}
\end{figure}

The second three-dimensional case study involves two initial semi-circular pits. The spatial domain is defined as $\mathcal{D} = [-80, 80] \times [-80, 80] \times [0, 40]\,\mathrm{\mu m}^3$, with the temporal domain spanning $\mathcal{T}=[0, 10]\,\mathrm{s}$. The two initial pits are positioned along the $x$-axis on the top boundary, each with a radius of $r = 10\,\mathrm{\mu m}$, separated by a center-to-center distance of $40\,\mathrm{\mu m}$. The initial conditions and boundary conditions for the phase field $\phi$ and concentration $c$ are depicted in Figure~\ref{fig:icbc-3d2pits}.

\begin{figure}[htbp]
    \centering
    \includegraphics[width=\textwidth]{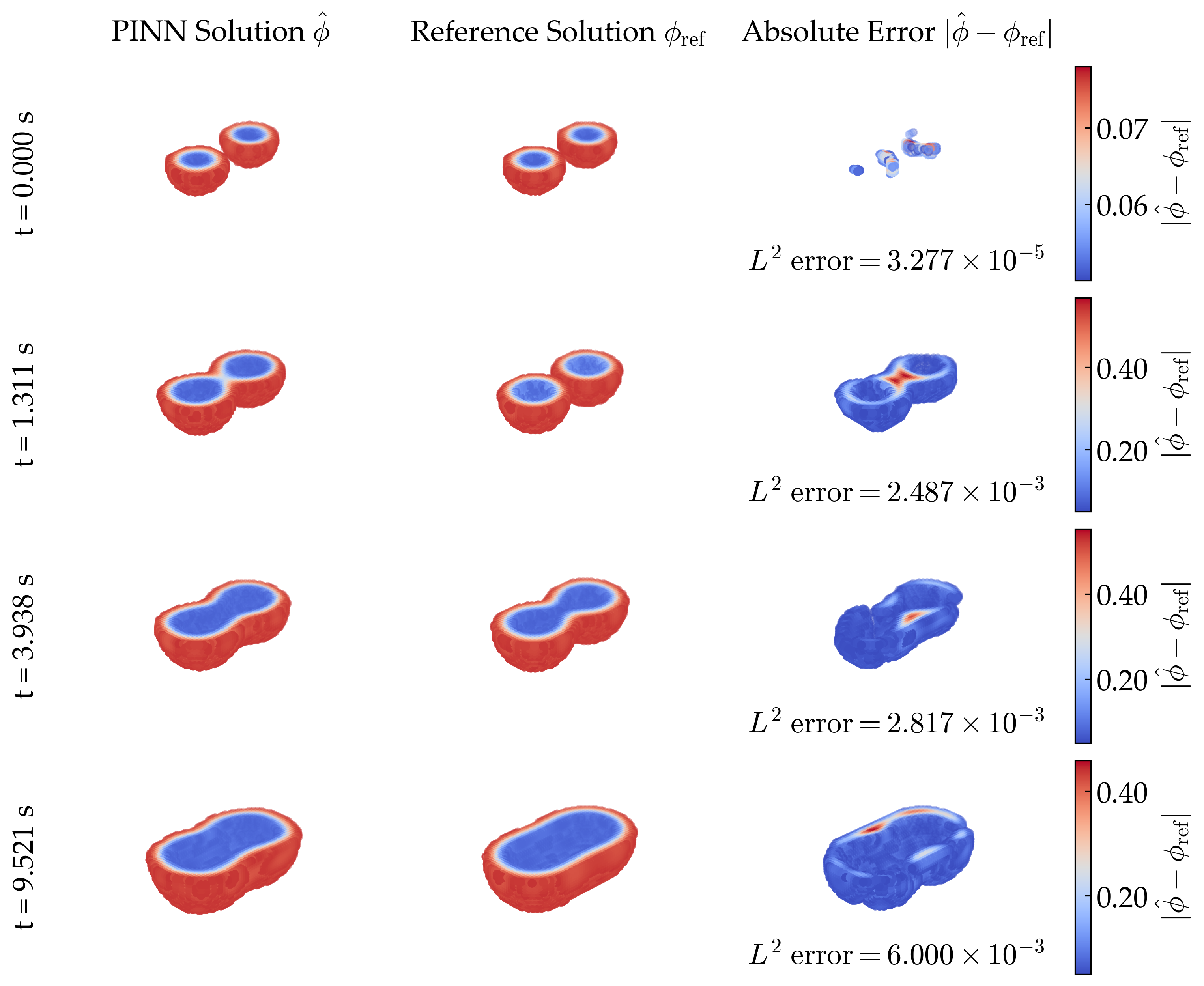}
    \caption{Three-dimensional semi-circular pitting corrosion with two pits interactions. Contours of phase field variable $\phi$ obtained from PINNs ($\hat \phi$), \texttt{FEniCS} ($\phi_\text{ref}$), and their absolute error ($|\hat \phi - \phi_\text{ref}|$). For visibility, the contours of $\hat \phi$ and $\phi_\text{ref}$ display only the interface region ($0.05 \leq \hat{\phi}, \phi_\text{ref} \leq 0.95$), while the error contours highlight the region where the absolute error exceeds $0.05$.}
    \label{fig:3d-2pits-phi-fields}
\end{figure}

The general sampling points are set to $N_g = 30\times 20\times 15\times 30 = 225,000$ for $x$, $y$, $z$, and $t$ dimensions, respectively, with all other hyperparameters and training configurations consistent with the default settings listed in Table~\ref{tab:hyperparameters}. Results of the PINN solution $\hat \phi$, the reference solution $\phi_\text{ref}$, and the absolute error $|\hat \phi - \phi_\text{ref}|$ at four representative time points are shown in Figure~\ref{fig:3d-2pits-phi-fields}. The overall accuracy of the Sharp-PINN model is still satisfactory, with the maximum absolute error of $6.000 \times 10^{-3}$ at $t = 9.521\,\mathrm{s}$. The complex pits coalescence process in three-dimensional space is accurately captured by the Sharp-PINN model. Additionally, the total runtime of the model is $18.23\,\mathrm{min}$, which is significantly more efficient than the finite element-based reference solution, which requires $197.23\,\mathrm{min}$ in total. This case study further demonstrates the accuracy and efficiency of the Sharp-PINN framework in solving complex three-dimensional phase field corrosion problems.

\begin{figure}[htbp]
    \centering
    \includegraphics[width=0.6\textwidth]{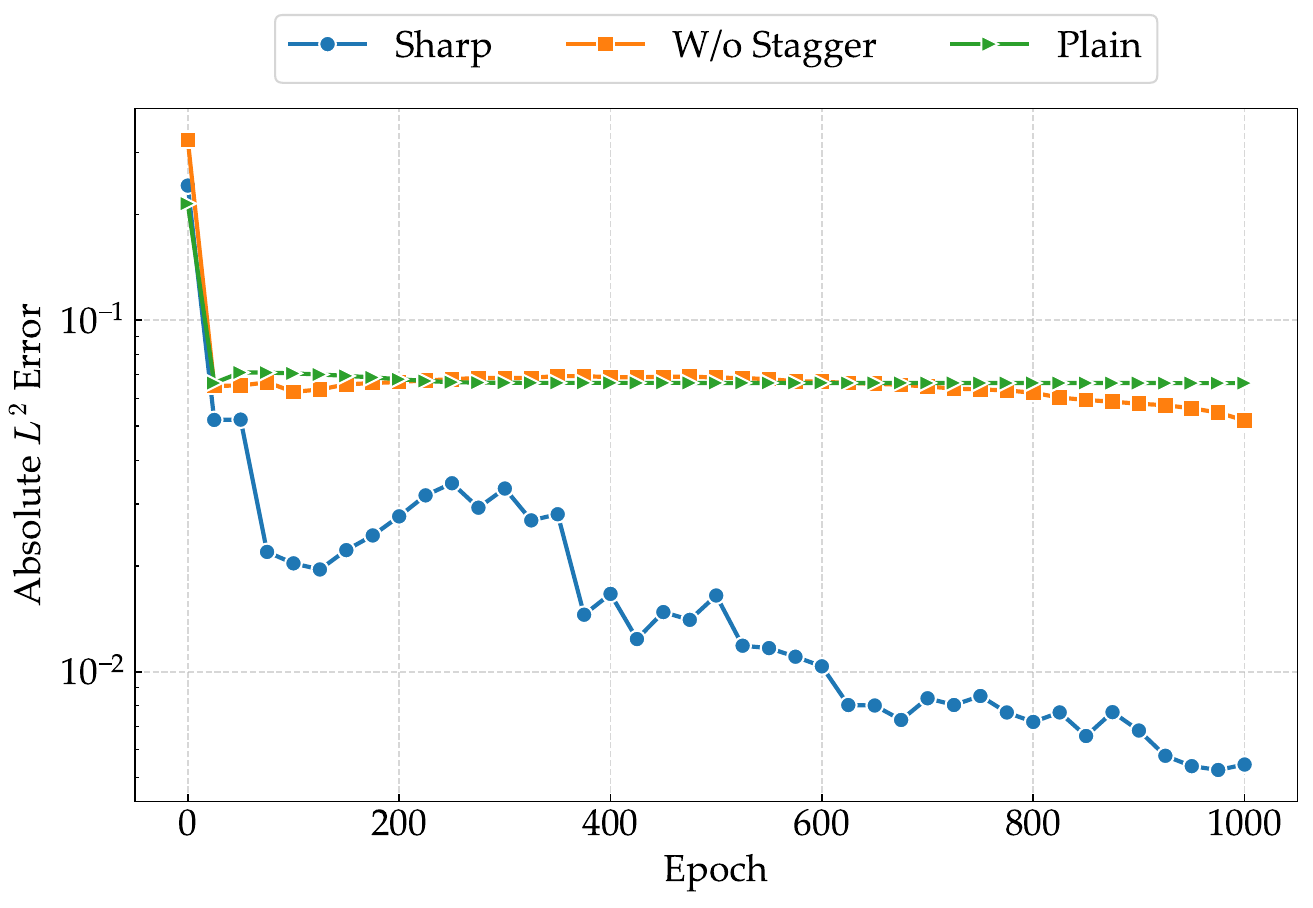}
    \caption{Three-dimensional corrosion with two-pits interactions. Convergence of absolute $L^2$ error of ablation studies. The abbreviations in the legend represent PINN models with different ablation configurations. Sharp: the proposed baseline model with all components enabled. W/o Stagger: PINN model without the staggered training scheme. Plain: the standard PINN model without any modifications.}
    \label{fig:3d-2pits-l2-error-history}
\end{figure}

To highlight the robustness of the staggered training scheme, we compare the Sharp-PINN model with all components enabled against two ablated models: one without the staggered training scheme and the other without any modifications. The ablation configurations and corresponding performance metrics are summarized in Table~\ref{tab:ablation-studies-3d2pits}. 
The convergence history of the absolute $L^2$ error for each ablation case is depicted in Figure~\ref{fig:3d-2pits-l2-error-history}. The results indicate that the Sharp-PINN model incorporating the staggered scheme achieves significantly faster convergence, yielding an absolute $L^2$ error of $1.345 \times 10^{-3}$, with a slightly increased yet justifiable computational cost compared to the ablated models.

\begin{table}[htbp]
    \centering
    \caption{Three-dimensional corrosion with two-pits interactions. Ablation configurations and performance metrics.}
    \label{tab:ablation-studies-3d2pits}
    \begin{tblr}{width=\textwidth,
        colspec={X[2,c]X[1,c]X[1,c]},
        rowspec={Q[m]Q[m]Q[m]Q[m]Q[m]Q[m]},
        row{1-2}={font=\bfseries}}
        \hline
        \SetCell[r=2]{c} Ablation Settings  & \SetCell[r=1,c=2]{c} Performance \\
        \cline{2-3}
        & Absolute $L^2$ error & Run time ($\text{min}$)  \\
        \hline
        Sharp-PINN model & $1.345\times10^{-3}$  & $18.40$ \\
        PINN without stagger & $5.188\times10^{-2}$ & $19.67$ \\
        Plain PINN model & $2.834\times10^{-2}$ & $8.68$ \\
        \hline
    \end{tblr}
\end{table}

\begin{figure}[ht]
    \centering
    \includegraphics[width=\textwidth]{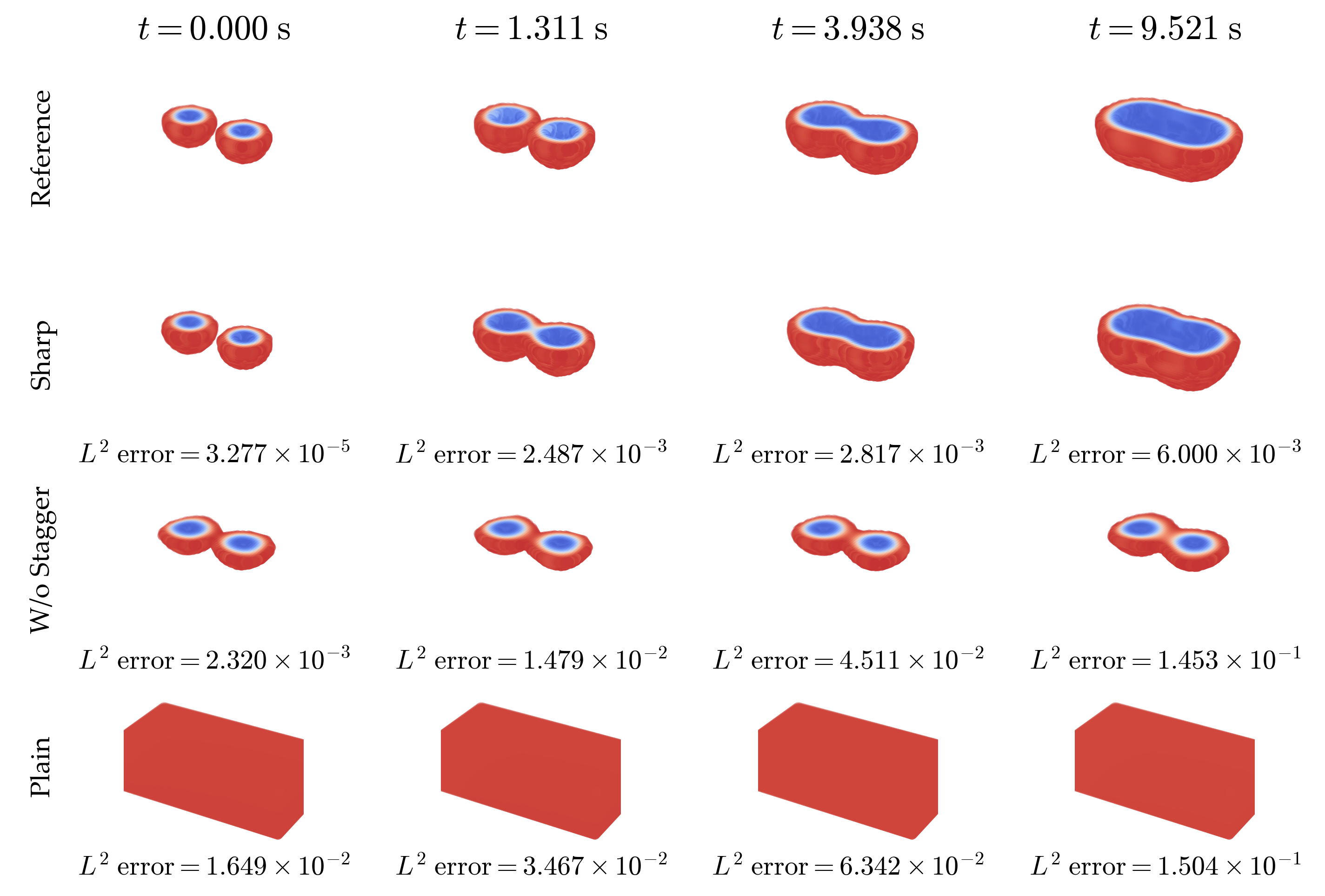}
    \caption{Three-dimensional corrosion with two-pits interactions. Comparison of the predicted phase field variable $\phi$ obtained from PINNs of different ablation settings. For visibility, the contours of $\hat \phi$ and $\phi_\text{ref}$ display only the interface region ($0.05 \leq \hat{\phi}, \phi_\text{ref} \leq 0.95$). The abbreviations in the left column are consistent with Figure~\ref{fig:3d-2pits-l2-error-history}.}
    \label{fig:3d-2pits-phi-fields-comparison-stagger}
\end{figure}

Figure~\ref{fig:3d-2pits-phi-fields-comparison-stagger} compares the predicted phase field $\phi$ across model variants in our ablation study. The results demonstrate that only our Sharp-PINN model can accurately capture the complex phase field evolution. In contrast, the ablated models exhibit severe performance degradation.
Specifically, the absence of the staggered training scheme leads to a constant spatial distribution of $\phi$ across all time points, indicating a failure to capture corrosion dynamics. Furthermore, the plain PINN model, without any enhancements, predicts a uniform and constant distribution of $\phi$ throughout the spatio-temporal domain, entirely failing to solve the phase field corrosion problem. These findings emphasizes the 
the critical importance of our methodological innovations described in Section \ref{sec:methodology}, particularly the staggered training strategy, for accurately modelling complex corrosion processes.

\section{Conclusions}
\label{sec:conclusion}

In this work, we propose Sharp-PINN, a novel physics-informed neural network framework with an efficient staggered training scheme for solving complex phase field corrosion problems. The strongly coupled phase field equations (AC and CH) are alternatively optimized in a staggered manner during the training process. Hard constraints are imposed to the output layer of the neural network, and explicitly formulate the composition of the concentration field. 

We extensively validate the proposed Sharp-PINN framework on a series of benchmark cases, including two- and three-dimensional complex corrosion problems with multiple initial pits. The results provide strong evidence of the framework's accuracy in capturing the complex phase field evolution. The thorough ablation studies further highlight the effectiveness and necessity of each component, and especially emphasizes the significant performance and efficiency gains achieved by the staggered training scheme. Most notably, in three-dimensional cases, the Sharp-PINN models achieve substantial reduction in computational cost compared to the finite element method. This suggests that the Sharp-PINN framework offers a promising alternative to traditional numerical methods for solving phase field corrosion problems.  

Building on this work, several exciting directions for future research can be identified. One promising avenue is to extend the framework to solve parametric versions of phase field equations by integrating physics-informed neural operators, potentially enabling real-time solutions across a range of parameter spaces. Another direction involves generalizing the staggered training scheme to other types of coupled PDEs, broadening its applicability to diverse physical systems. Additionally, investigating more complex and realistic scenarios within the phase field corrosion domain, such as incorporating multi-physical coupling effects or considering multi-scale corrosion processes, represents an exciting opportunity to further advance the capability of the Sharp-PINNs framework. These developments have the potential to significantly expand the scope and impact of PINNs in scientific computing.

\section*{CRediT authorship contribution statement}
\textbf{Nanxi Chen:} Software, Investigation, Formal analysis, Writing - original draft.
\textbf{Chuanjie Cui:} Conceptualization, Validation, Writing - Review \& Editing, Project administration.
\textbf{Rujin Ma:} Conceptualization, Resources, Supervision, Funding acquisition.
\textbf{Airong Chen:} Conceptualization, Resources, Supervision, Funding acquisition.
\textbf{Sifan Wang:} Methodology, Investigation, Validation, Writing - Review \& Editing.

\section*{Declaration of Competing Interest}
The authors declare that they have no known competing financial interests or personal relationships that could have appeared to influence the work reported in this paper.

\section*{Acknowledgments}
The authors acknowledge financial support from the National Natural Science Foundation of China (grants No.52178153, No.52478199, and No.52238005). S.W. thanks the support from Yale Institute for Foundations of Data Science and NVIDIA Academic Grant Program.

\section*{Data availability}
Data and code used in this paper are made freely available at \url{https://github.com/NanxiiChen/sharp-pinns}.  Detailed annotations of the code are also provided.

\bibliographystyle{elsarticle-num}
\biboptions{numbers,sort&compress}
\bibliography{bibfile-cnx,bibfile-ccj}

\clearpage
\appendix
\setcounter{table}{0}
\setcounter{figure}{0}

\section{Hyperparameters}

\begin{table}[htbp]
    \centering
    \caption{Physical parameters used in the phase field corrosion model (SI units).}
    \label{tab:parameters}
    \begin{tblr}{width=\textwidth,colspec={X[1,c]X[5,c]X[1,c]},row{1}={font=\bfseries}}
        \hline
        Notation       & Description & Value \\
        \hline
        $\alpha_\phi$   & Gradient energy coefficient                               & $1.03\times10^{-4}$  \\
        $w_\phi$        & Height of the double well potential                       & $1.76\times 10^7$    \\
        $\ell$          & Interface thickness                                       & $1.0\times 10^{-5}$   \\
        $L$             & Mobility parameter                                        & $2.0$                \\
        $M$             & Diffusivity parameter                                     & $7.94\times10^{-18}$ \\
        $\mathcal{A}$   & Free energy density-related parameter                     & $5.35\times 10^7$    \\
        $c_\mathrm{Se}$ & Normalised equilibrium concentration for the solid phase  & $1.0$                \\
        $c_\mathrm{Le}$ & Normalised equilibrium concentration for the liquid phase & $0.036$              \\
        \hline
    \end{tblr}
\end{table}

\begin{table}[htbp]
    \centering
    \caption{Hyperparameters used in the Sharp-PINNs framework.}\label{tab:hyperparameters}
    \begin{threeparttable}
    \begin{tblr}{
        width=\textwidth,
        colspec={X[1,c]X[3.5,c]X[1,c]X[1.5,c]},
        row{1}={font=\bfseries}}
        \hline
        Group & Description & Notation & Value \\
        \hline
        \SetCell[r=4]{m} {Network\\architecture} 
        & Fourier feature Dimension & $m_f$ & $64$\\
        & Fourier feature Frequencies & $\sigma_x$, $\sigma_t$& $2.0$,$0.4$ \\
        & Hidden layer width & $m_w$ &  $128$ \\
        & Hidden layer depth & $m_h$ &  $6$ \\
        \hline
        \SetCell[r=3]{m} Sampling 
        & Number of general collocation points & $N_g$ 
        & $40\times20\times30$ \\
        & Number of boundary points & $N_b$&  $500$ \\
        & Number of initial points & $N_i$&   $800$ \\
        \hline
        \SetCell[r=3]{m} {Training\\process} 
        & Staggering period & $S_s$ &  $25$\\
        & Weighting update rate & $\alpha_w$ & $0.5$\\
        & Initial learning rate & $\eta$ & $5\times10^{-4}$\\
        \hline
    \end{tblr}
    \begin{tablenotes}
        \footnotesize
        \item Note: The general collocation points are randomly sampled along each axis and then combined to form the spatial-temporal coordinates, formulated in $N_g = N_{g, x}\times N_{g, y}\times N_{g, t}$ for 2D cases and $N_g = N_{g, x}\times N_{g, y}\times N_{g, z}\times N_{g, t}$ for 3D cases. Here, we list only the number of collocation points for 2D cases.
      \end{tablenotes}
\end{threeparttable}
\end{table}

\section{Grad-norm-based loss balancing scheme}
\label{sec:loss-balancing}

The loss function under the staggered training scheme is defined as the weighted sum of the loss terms associated with the PDE (AC or CH), boundary conditions, and initial conditions, as shown in Eq.~\eqref{eq:loss-stage1}~and~\eqref{eq:loss-stage2}. A grad norm weighting scheme \cite{chenGradNormGradientNormalization2018,wangExpertsGuideTraining2023} is employed to balance these loss terms. The weights $w_\text{AC}$, $w_\text{CH}$, $w_\text{BC}$, and $w_\text{IC}$ are determined based on the gradients of the loss terms, ensuring that the network is equally trained on all components of the loss function and preventing the optimization process from being dominated by a single term. For a training step $s$, the weights of each loss term are calculated as:
\begin{subequations}
    \begin{gather}
        \hat w_j^{(s)} = \frac{\displaystyle\sum_{j\in\mathcal{J}} \| \nabla_{\bm\theta}\mathcal{L}_j^{(s)} \|}{\| \nabla_{\bm\theta}\mathcal{L}_j^{(s)} \|},
        \quad s \geq 1, \quad \forall j\in\mathcal{J}, \\
        w_j^{(s)} = \alpha_w \cdot \hat w_j^{(s-1)} + (1-\alpha_w) \cdot w_j^{(s)} \\
        w_j^{(0)} = 1,
    \end{gather}%
\end{subequations}
where $\|\cdot\|$ denotes the $L^2$ norm, $\alpha_w$ is a hyperparameter that controls the weighting update rate (given in Table~\ref{tab:hyperparameters}). $\mathcal{J}$ represents the name set of loss terms. Specifically, for staggered training stage 1, $\mathcal{J} = \{\text{AC}, \text{BC}, \text{IC}\}$, and for stage 2, $\mathcal{J} = \{\text{CH}, \text{BC}, \text{IC}\}$.

\section{Random Fourier feature for the input embedding}
\label{sec:random-fourier-feature-embedding}

Spectral bias \cite{wangEigenvectorBiasFourier2021,wangWhenWhyPINNs2022} is a well-documented limitation of PINNs, which hinders their ability to accurately approximate functions with high-frequency components. In phase field corrosion problems, the phase field, $\phi$, and concentration, $c$, exhibit sharp transitions within the diffuse interface region, while remain relatively smooth outside this region. The stark contrast in the spatial frequency characteristics of the solution field presents great challenges to PINNs to learn the target solution effectively.

Random Fourier feature embedding  \cite{tancikFourierFeaturesLet2020} is a simple yet efficient method to mitigate spectral bias in PINNs, which embeds input coordinates into some high frequency signals in the latent space. Specifically, a random Fourier mapping $\mathcal{F}$ is defined as:
\begin{equation}
    \mathcal{F} (\bm v) = \left[
        \begin{matrix}
            \cos (\bm B \bm v) \\ \sin (\bm B \bm v)
        \end{matrix}
    \right],
\end{equation}%
where $\bm B \in \mathbb{R}^{m_f\times d}$ is a matrix whose entries are sampled from a Gaussian distribution $\mathcal{N}(0, \sigma^2)$. Here $m_f$ represents the number of Fourier features, $d$ is the dimension of the input space, and $\sigma$ is a hyperparameter that controls the frequency range of the Fourier features.

In this work, we apply the random Fourier feature embedding separately to the spatial coordinates $\bm x$ and the temporal coordinate $t$, and concatenate the resulting features as the input to the neural network. This process can be expressed as: 
\begin{equation}
    \bm z = \mathcal{F}(\bm x, t) = \mathcal{F}_x(\bm x) \oplus \mathcal{F}_t(t) = \left[
        \begin{matrix}
            \cos (\bm B_x \bm x) \\ \sin (\bm B_x \bm x) \\ \cos (\bm B_t t) \\ \sin (\bm B_t t)
        \end{matrix}
    \right],
\end{equation}%
where $\bm B_x$ and $\bm B_t$ are random matrices corresponding to the spatial and temporal Fourier features, respectively. These matrices are independently sampled with hyperparameters $\sigma_x$ and $\sigma_t$, whose values are provided in Table~\ref{tab:hyperparameters}. The concatenated feature vector $\bm z$, with a total of $4m_f$ dimensions, is then passed to the neural network to approximate the solution field.

\section{Modified MLP architecture for the backbone network}
\label{sec:modified-mlp-architecture}

\begin{figure}[H]
    \centering
    \includegraphics[width=\textwidth]{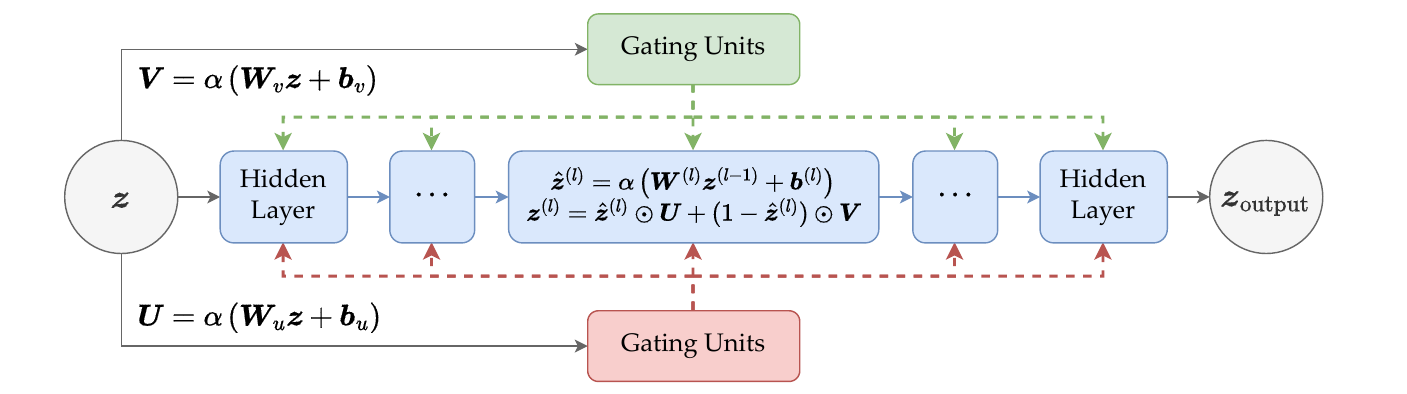}
    \caption{Schematic of the modified MLP architecture.}
    \label{fig:modified-mlp}
\end{figure}

We also employ a modified MLP architecture $\mathcal{M}$ to efficiently learn solutions of nonlinear PDEs.
This architecture introduces gating mechanisms and residual connections to the standard MLP, enhancing the network's capability of capturing complex PDE solutions. A schematic of the modified MLP architecture is shown in Figure~\ref{fig:modified-mlp}.
Firstly, two fully connected layers serve as the gating units to modulate the information flow between the hidden layers. For the input feature $\bm z$, the gating vectors $\bm U$ and $\bm V$ are computed as:
\begin{subequations}
    \begin{align}
        \bm U &= \alpha\left(\bm W_u \bm z + \bm b_u\right), \\
        \bm V &= \alpha\left(\bm W_v \bm z + \bm b_v\right),
    \end{align}
\end{subequations}%
where \(\alpha\) denotes the activation function. Then, for the $l$-th (with $l = 1, 2, \ldots, L$) hidden layer, the hidden state $\bm z^{(l)}$ is updated using the weighted sum of the previous hidden state $\bm z^{(l-1)}$ and the gating information:
\begin{subequations}
    \begin{gather}
        \hat{\bm{z}}^{(l)} = \alpha \left( \bm W^{(l)} \bm z^{(l-1)} + \bm b^{(l)} \right),\\
        \bm z^{(l)} = \hat {\bm z}^{(l)} \odot \bm U + \left( 1 - \hat {\bm z}^{(l)}\right) \odot \bm V,
    \end{gather}
\end{subequations}%
where $\odot$ denotes element-wise multiplication. The final output of the network is computed by passing the last hidden state \(\bm z^{(L)}\) through a linear layer:
\begin{equation}
    \bm z_\text{output} = \bm W^{(L+1)} \bm z^{(L)} + \bm b^{(L+1)},
\end{equation}%
Here, $\bm W_u, \bm W_v, \{\bm W^{(l)}\}_{l=1}^{L+1}$ and $\bm b_u, \bm b_v, \{\bm b^{(l)}\}_{l=1}^{L+1}$ are the weights and biases of each layer, respectively.

\section{Implementation details for applying PINNs to phase field corrosion model}
\label{sec:implementation-details}

\subsection{Non-dimensionalization for the governing equations}
\label{sec:non-dimensionalization}

The phase field method is commonly applied to mesoscopic systems, where spatial scales are typically on the order of micrometers, and time scales can extend from seconds to hours. This wide range of input values for spatial and temporal coordinates poses significant challenges for a neural network to learn the solution effectively \cite{xuPreprocessingPhysicsinformedNeural2024}. In order to limit the input range to a reasonable scale and improve the numerical stability of the neural network. we shall conduct non-dimensionalization of the governing equations. Firstly, we define the characteristic length $L_c$ and time $T_c$ as the domain size and the characteristic diffusion time, respectively. The non-dimensional variables are defined as:
\begin{subequations}
    \begin{gather}
        \bm x^* = \frac{\bm x}{L_c}, \\
        t^* = \frac{t}{T_c}.
    \end{gather}\label{eq:non-dimensionalization}
\end{subequations}%
Then four non-dimensional parameters are defined as:
\begin{equation}
    N_\text{CH} = 2\mathcal{A}M\frac{T_c}{L_c^2}, \quad N_\text{AC,1} = 2\mathcal{A}LT_c, \quad N_\text{AC,2} = Lw_\phi T_c, \quad N_\text{AC,3} = L\alpha_\phi\frac{T_c}{L_c^2}.
\end{equation}%
Substituting the non-dimensional variables and parameters into Eqs.~\eqref{eq:governing-equations-corrosion}, the non-dimensional governing equations can be rewritten as:
\begin{subequations}
    \begin{gather}
        \frac{\partial c}{\partial t^*}- N_\text{CH} \Delta c + N_\text{CH}\left(c_{\mathrm{Se}}-c_{\mathrm{Le}}\right) \Delta^* h\left(\phi\right) =0,  \label{eq:ch-non-dim-parameters} \\
        \frac{\partial \phi}{\partial t^*} - N_\text{AC,1}\left[c-h(\phi)\left(c_{\mathrm{Se}}-c_{\mathrm{Le}}\right)-c_{\mathrm{Le}}\right]\left(c_{\mathrm{Se}}-c_{\mathrm{Le}}\right) h^{\prime}(\phi) + N_\text{AC,2} g^{\prime}(\phi) - N_\text{AC,3}\Delta^* \phi =0.
        \label{eq:ac-non-dim-parameters}
    \end{gather}
    \label{eq:governing-equations-non-dimensional-parameters}
\end{subequations}%
In this study, the characteristic length $L_c$ is set to $1.0\times 10^{-4} \,\mathrm{m}$ and the characteristic time $T_c$ is set to $10.0\,\mathrm{s}$.

\subsection{Initial condition for the diffused interface}
\label{sec:initial-condition-diffused-interface}

In the phase field corrosion model, the evolution process typically involves the separation of solid and liquid phases, characterized by a diffuse interface with finite width. To ensure the physical consistency of the model, the initial condition must not only define the position of the initial interface but also specify the accurate profiles of the phase field and concentration field \cite{chenPFPINNsPhysicsinformedNeural2025,chenPCPINNsPhysicsinformedNeural2024a}. An appropriate function is required to describe the smooth transition within the diffuse interface region. Here, we follow the definition of the KKS model \cite{kimPhasefieldModelBinary1999} to construct the initial condition for the diffuse interface. The regularised topology of the phase field profile is given by
\begin{equation}
    \phi_0(x_d) = \frac{1}{2}\left[
        1 - \tanh\left(
        \frac{\sqrt{w_\phi}}{\sqrt{2\alpha_\phi}}x_d
        \right)
        \right], \label{eq:phixd}
\end{equation}
where $x_d$ is the distance from the initial interface, and $\phi_0=1$ (solid) is assumed as $x_d \to -\infty$. Accordingly, the initially diffused concentration $c_0(x_d)$ can be defined as
\begin{equation}
    c_0(x_d) = h\left[ \phi_0(x_d) \right]c_\mathrm{Se}. \label{eq:cdiffuse}
\end{equation}%

Additionally, considering the gradients of field variables are negligible outside the initial interface where the evolution begins, global sampling is applied across the entire spatial domain with a reduced density. While in the vicinity of the initial interface region, normally 1-2 times the interface thickness, a denser sampling is required.

\section{FEM implementation for phase field corrosion model}
\label{sec:fem-implementation}

This Appendix provides the \texttt{FEniCS}-based FEM implementation for the phase field corrosion model. The strong form of the governing equations is given by Eqs.~\eqref{eq:governing-equations-corrosion}. To derive the weak form, we define the test functions $v_c$ and $v_\phi$ for the concentration and phase field equations, respectively. The weak form of the governing equations can be expressed as:
\begin{subequations}
    \begin{gather}
        \int_{\Omega} \frac{\partial c}{\partial t} v_c \, \mathrm{d}\Omega 
    + \int_{\Omega} 2\mathcal{A}M \nabla c \cdot \nabla v_c \, \mathrm{d}\Omega 
    - \int_{\Omega} 2\mathcal{A}M \left(c_{\mathrm{Se}} - c_{\mathrm{Le}}\right) \nabla h(\phi) \cdot \nabla v_c \, \mathrm{d}\Omega = 0.\label{eq:cahn-hilliard-weak-form} \\
    \begin{aligned}
        \int_{\Omega} \frac{\partial \phi}{\partial t} v_\phi \, \mathrm{d}\Omega 
        - \int_{\Omega} 2\mathcal{A}L \left[c - h(\phi)\left(c_{\mathrm{Se}} - c_{\mathrm{Le}}\right) - c_{\mathrm{Le}}\right] \left(c_{\mathrm{Se}} - c_{\mathrm{Le}}\right) h^{\prime}(\phi) v_\phi \, \mathrm{d}\Omega&\\
        + \int_{\Omega} L w_\phi g^{\prime}(\phi) v_\phi \, \mathrm{d}\Omega 
        + \int_{\Omega} L \alpha_\phi \nabla \phi \cdot \nabla v_\phi \, \mathrm{d}\Omega &= 0.
    \end{aligned}\label{eq:allen-cahn-weak-form}
    \end{gather}
\end{subequations}%
where $\Omega$ denotes the spatial domain and the remaining notations are consistent with those provided in Table~\ref{tab:parameters}.

Combining the weak forms of AC and CH equations, along wiht the initial and boundary conditions, the phase field corrosion problem can be solved using FEM in \texttt{FEniCS}, which serves as the reference solution in Section~\ref{sec:results-and-discussion}. Considering the interface thickness of $10 \,\mathrm{\mu m}$, the spatial domain is discretized into a mesh size of approximately $2\,\mathrm{\mu m}$. Since the corrosion process slows down as the interface evolves, we set a tiny time step of $0.001 \,\mathrm{s}$ at the beginning and adaptively increase as the simulation progresses. Algorithm~\ref{alg:fenics} provides a general outline of the FEM implementation for the phase field corrosion model.

\begin{algorithm}[H]
    \caption{FEM implementation for phase field corrosion model using \texttt{FEniCS}.}
    \label{alg:fenics}
        Define the mesh for the computational domain $\Omega$, the trial and test function spaces for $c$ and $\phi$, and the model parameters according to Table~\ref{tab:parameters}\;
    
        Define the boundary conditions and initial conditions\;
        Formulate the weak forms of governing equations according to Eqs. \eqref{eq:cahn-hilliard-weak-form} and \eqref{eq:allen-cahn-weak-form}\;

        Initialize time $t = 0$ and time step $\Delta t$\;
        \While{
            $t \leq t_{\text{end}}$
        }{
            Try solving the coupled nonlinear system\;
            \eIf{
                Converged
            }{
                Update $c$ and $\phi$ according to the solution\;
                Step forward $t \leftarrow t + \Delta t$\;
                \If{
                    Iteration $< 5$
                }{
                    Increase time step $\Delta t \leftarrow \Delta t \times 1.5$\;
                }
                
            }{
                Decrease time step $\Delta t \leftarrow \Delta t / 2$\;
                Retry solving\;
                \If {
                    $\Delta t < \Delta t_{\text{min}}$
                }{
                    Break\;
                }
            }
        }
\end{algorithm}%

Here, $t_{\text{end}}$ denotes the final time, and $\Delta t_{\text{min}}$ is the tolerance for the minimum time step. The time step is doubled if the solution converges within the first five iterations.






\end{document}